%% file: main.tex
\documentclass[twoside,journal]{IEEEtran}

\newtheorem{theorem}{Theorem}[section]
\newtheorem{prop}[theorem]{Proposition}
\newtheorem{assumption}[theorem]{Assumption}
\newtheorem{problem}{Problem}
\newtheorem{definition}[theorem]{Definition}
\newtheorem{rem}[theorem]{Remark}
\newtheorem{ex}[theorem]{Example}

\IEEEoverridecommandlockouts
\usepackage{graphicx}

\usepackage{epsfig} 
\usepackage{epsfig} 
\usepackage{amsmath}
\usepackage{amssymb}
\usepackage{amssymb}
\usepackage{epstopdf}
\usepackage{cite}
\usepackage{savesym}
\savesymbol{AND}
\usepackage{algorithm}
\usepackage{algpseudocode}
\usepackage{multirow}
\usepackage{rotating}
\usepackage{subfigure}
\usepackage{color}
\usepackage{mysymbol}
\usepackage{url}	

\begin{document}
 
\title{Sample-Efficient Reinforcement Learning with Temporal Logic Objectives: Leveraging the Task Specification to Guide Exploration}

\author{Yiannis~Kantaros,~\IEEEmembership{Member,~IEEE}, Jun Wang  ~\IEEEmembership{Student Member,~IEEE}
\thanks{Yiannis Kantaros ($\text{ioannisk}$@wustl.edu) and Jun Wang ($\text{junw}$@wustl.edu) are with the Department of Electrical and Systems Engineering, Washington University in St. Louis, St. Louis, MO, 63130, USA. This work was supported in part by the NSF award CNS $\#2231257$.}.}

\maketitle
\begin{abstract}
This paper addresses the problem of learning optimal control policies for systems with uncertain dynamics and high-level control objectives specified as Linear Temporal Logic (LTL) formulas. Uncertainty is considered in the workspace structure and the outcomes of control decisions giving rise to an unknown Markov Decision Process (MDP). Existing reinforcement learning (RL) algorithms for LTL tasks typically rely on exploring a product MDP state-space uniformly (using e.g., an $\epsilon$-greedy policy) compromising sample-efficiency. This issue becomes more pronounced as the rewards get sparser and the MDP size or the task complexity increase. In this paper, we propose an accelerated RL algorithm that can learn control policies significantly faster than competitive approaches. Its sample-efficiency relies on a novel task-driven exploration strategy that biases exploration towards directions that may contribute to task satisfaction. We provide theoretical analysis and extensive comparative experiments demonstrating the sample-efficiency of the proposed method. The benefit of our method becomes more evident as the task complexity or the MDP size increases. 
\end{abstract}
\vspace{-0.1cm}

\vspace{-0.4cm}
\section{Introduction} \label{sec:Intro}
\vspace{-0.1cm}

\input{files/IntroV2}
\vspace{-0.2cm}
\section{Problem Definition} \label{sec:PF}
\vspace{-0.1cm}
\input{files/PF_short_v3}

\vspace{-0.2cm}
\section{Accelerated Reinforcement Learning for Temporal Logic Control}
%
\label{sec:planning}

\input{files/alg2}

\vspace{-0.3cm}
\section{Algorithm Analysis}\label{sec:theory}
\vspace{-0.1cm}
\input{files/theory}

\vspace{-0.2cm}
\section{Numerical Experiments} \label{sec:Sim}
\vspace{-0.1cm}
\input{files/sim}
\section{Conclusions}\label{sec:conclusion}
In this paper, we proposed a new accelerated reinforcement learning (RL) for temporal logic control objectives. The proposed RL method relies on new control policy, called $(\epsilon,\delta)$-greedy, that prioritizes exploration in the vicinity of task-related regions. This results in enhanced sample-efficiency as supported by theoretical results and comparative experiments.
Our future work will focus on enhancing scalability by using function approximations (e.g., neural networks). 
%
 
\appendices
\vspace{-0.3cm}
\section{Extensions: Biased Exploration over LDBA}\label{app:ldba}
\vspace{-0.1cm}
\input{files/extensionsLDBA}

\vspace{-0.5cm}
\section{Proof for Results of Section \ref{sec:myopic}}\label{sec:proof1} 
\vspace{-0.1cm}
\input{files/proofs1}

\vspace{-0.3cm}
\section{Proof of Results of Section \ref{sec:nonmyopicBias}}\label{sec:proof2} 
\input{files/proofs2_v3} 
\vspace{-0.3cm}

\section{Decay Rates in Numerical Simulations}\label{sec:decayRates} 
\input{files/decay}


\bibliographystyle{IEEEtran}
\bibliography{YK_bib.bib}

\vspace{-3cm}
\begin{IEEEbiography}[]
{Yiannis Kantaros}(S'14-M'18) is an Assistant Professor in the Department of Electrical and Systems Engineering, Washington University in St. Louis (WashU), St. Louis, MO, USA. He received the Diploma in Electrical and Computer Engineering in 2012 from the University of Patras, Patras, Greece. He also received the M.Sc. and the Ph.D.  degrees in mechanical engineering from Duke University, Durham, NC, in 2017 and 2018, respectively. Prior to joining WashU, he was a postdoctoral associate in the Department of Computer and Information Science, University of Pennsylvania, Philadelphia, PA. His current research interests include machine learning, distributed control and optimization, and formal methods with applications in robotics. He received the Best Student Paper Award at the IEEE Global Conference on Signal and Information Processing (GlobalSIP) in 2014, a Best Multi-Robot Systems Paper Award, Finalist, at the IEEE International Conference in Robotics and Automation (ICRA) in 2024, the 2017-18 Outstanding Dissertation Research Award from the Department of Mechanical Engineering and Materials Science, Duke University, and a 2024 NSF CAREER Award.
\end{IEEEbiography}
\vspace{-3cm}
\begin{IEEEbiography}[]
{Jun Wang}(S'22) is a PhD candidate in the Department of Electrical and Systems Engineering at Washington University in St. Louis. He received his B.Eng. degree in Software Engineering from Sun Yat-Sen University in 2019 and his MSE degree in Robotics from the University of Pennsylvania in 2021. His research interests include robotics, machine learning, and control theory.
\end{IEEEbiography}

\end{document}

%% file: files/IntroV2.tex
Reinforcement learning (RL) has been successfully applied to synthesize control policies for systems with highly nonlinear, stochastic or unknown dynamics and complex tasks \cite{kiran2021deep}. 
Typically, in RL, control objectives are specified as reward functions. 
However, specifying reward-based objectives can be highly non-intuitive, especially for complex tasks, while poorly designed rewards can significantly compromise system performance \cite{dewey2014reinforcement}. To address this challenge, Linear Temporal logic (LTL) has recently been employed to specify 
tasks 
that would have been very hard to define using Markovian rewards \cite{baier2008principles}; e.g., consider a navigation task requiring to visit regions of interest in a specific order.

Several model-free RL methods with LTL-encoded tasks have been proposed recently; see e.g., \cite{wang2015temporal,hahn,gao2019reduced,bouton2019reinforcement,hasanbeig2019reinforcement,bozkurt2020control,cai2020learning,lavaei2020formal,jothimurugan2021compositional,hasanbeig2022lcrl,hasanbeig2023certified,bozkurt2024learning}. Common in the majority of these works is that they explore \textit{randomly} a product state space that grows exponentially as the size of the MDP and/or the complexity of the assigned temporal logic task increase. This results in sample inefficiency and slow training/learning process. This issue becomes more pronounced by the fact that LTL specifications are converted into sparse rewards in order to synthesize control policies with probabilistic satisfaction guarantees \cite{bozkurt2020control,hasanbeig2023certified,xuan2024uniqueness}.
Sample inefficiency is a well-known limitation in RL, whether control objectives are specified using reward functions directly or LTL. To address this limitation, reward engineering approaches have been proposed augmenting the reward signal \cite{hasanbeig2021deepsynth, icarte2018using, wen2020efficiency, toro2022reward, cai2022learning, balakrishnan2023model, pathak2017curiosity}. 
Such methods often require a user to \textit{manually} decompose the global task into sub-tasks, followed by assigning additional rewards to these intermediate sub-tasks. Nevertheless, this may result in sub-optimal control policies concerning the original task \cite{zhai2022computational}, while their efficiency highly depends on the task decomposition (i.e., the density of the rewards) \cite{cai2022overcoming}. Also, augmenting the reward signal for temporal logic tasks may compromise the probabilistic correctness of the synthesized controllers \cite{bozkurt2020control}.
To alleviate these limitations, intelligent exploration strategies have been proposed, such as Boltzmann/softmax \cite{kaelbling1996reinforcement,cesa2017boltzmann} and upper confidence bound (UCB) \cite{chen2017ucb} that do not require knowledge or modification of the rewards; a recent survey is available in \cite{amin2021survey}. Their sample-efficiency relies on guiding exploration using a continuously learned value function (e.g., Boltzmann) which, however, can be inaccurate in early training episodes. Alternatively, they rely on how many times a state-action pair has been visited (e.g., UCB), which might not always guide exploration towards directions contributing to task satisfaction. 

\textcolor{black}{Another approach to enhance sample-efficiency is through model-based methods \cite{fu2014probably, brazdil2014verification}. These works continuously learn an unknown Markov Decision Process (MDP), modeling the system, that is composed with automaton representations of LTL tasks. This gives rise to a product MDP (PMDP). Then, approximately optimal policies are constructed for the PMDP in a finite number of iterations. However, saving the associated data structures for the PMDP results in excessive memory requirements. Also, the quality of the generated policy critically depends on the accuracy of the learned PMDP. 
Finally, model-based methods require the computation of accepting maximum end components (AMECs) of PMDPs that has a quadratic time complexity in the PMDP size. This computation is avoided in related model-free methods; see e.g., \cite{gao2019reduced}.}
\textcolor{black}{In this paper, we propose a novel  approach to enhance the sample-efficiency of model-free RL methods. 
Unlike the aforementioned works, the key idea to improve sample efficiency is to leverage the (known) task specification in order to extract promising directions for exploration that contribute to mission progress.} We consider robots modeled as unknown MDPs with discrete state and action spaces, modeling uncertainty in the workspace and in the outcome of control decisions, and high-level LTL-encoded control objectives. 
%
The proposed algorithm relies on the following three steps. First, the LTL formula is converted into a Deterministic Rabin Automaton (DRA). Second, similar to \cite{gao2019reduced}, the product between the MDP and the DRA is constructed on-the-fly giving rise to a  PMDP over which rewards are assigned based on the DRA acceptance condition. We note that the PMDP is not explicitly constructed/strored in our approach. The first two steps are common in related model-free algorithms. Third, a new RL method is applied over the PMDP to learn policies that maximize the expected accumulated reward capturing the satisfaction probability. 
The proposed RL algorithm relies on a new stochastic policy, called $(\epsilon,\delta)-$ greedy policy, that exploits the DRA representation of the LTL formula to bias exploration towards directions that may contribute to task satisfaction. Particularly, according to the proposed policy, the greedy action is selected with probability $1-\epsilon$ (exploitation phase) while exploration is triggered with probability $\epsilon$, as in the $\epsilon$-greedy policy. Unlike the $\epsilon$-greedy policy, when exploration is enabled, either a random or a biased action is selected probabilistically (determined by $\delta$ parameters), where the latter action guides the system towards directions that will most likely result in mission progress.
For instance, consider a simple scenario where a robot with uncertain/unknown dynamics is required to eventually safely reach a region of interest. In this case, intuitively, exploration in the vicinity of the shortest dynamically feasible path (that is initially unknown but it is continuously learned)  connecting the current robot position to the desired region should be prioritized to accelerate control design. \textcolor{black}{We emphasize that the proposed task-driven exploration strategy does not require knowledge or modification of the reward structure. As a result, it can be coupled with sparse rewards, as e.g., in \cite{bozkurt2020control,hasanbeig2022lcrl}, resulting in probabilistically correct control policies as well as with augmented rewards, as e.g., in \cite{toro2022reward,cai2022overcoming,balakrishnan2023model}, to further accelerate the learning phase. 
} 


Our approach is inspired by transfer learning algorithms  that leverage external teacher policies for `similar' tasks to bias exploration \cite{fernandez2010probabilistic}.  To design a biased exploration strategy, in the absence of external policies, we build upon \cite{kantaros2020stylus,kantaros2022perception} that propose a biased sampling-based strategy to synthesize temporal logic controllers for large-scale, but \textit{deterministic}, multi-robot systems. Particularly, computation of the biased action requires (i) a distance function over the DRA state space,  similarly constructed as in \cite{kantaros2020stylus,kantaros2022perception,ding2014ltl,lacerda2015optimal}, to measure how far the system is from satisfying the assigned LTL task, and (ii) a continuously learned MDP model. The latter renders the proposed exploration strategy model-based. \textcolor{black}{Thus, we would like to emphasize the following key differences with respect to related model-based RL methods discussed earlier. First, unlike existing model-based algorithms, the proposed method does not learn/store the PMDP model to compute the optimal policy. Instead, it learns only the MDP modeling the system, making it more memory efficient. Second, the quality of the learned policy  is not contingent on the quality of the learned MDP model, distinguishing it from model-based methods.
This is because our approach utilizes the MDP model solely for designing the biased action and, in fact, as it will be discussed in Section \ref{sec:biasedExpl}, does not even require learning \textit{all} MDP transition probabilities accurately. 
This is also supported by our numerical experiments where we empirically demonstrate sample efficiency of the proposed method against model inaccuracies.
%
}
We provide comparative experiments demonstrating that the proposed learning algorithm outperforms in terms of sample-efficiency model-free RL methods that employ random (e.g., \cite{gao2019reduced,hasanbeig2019reinforcement, bozkurt2020control}), Boltzmann, and UCB exploration. The benefit of our approach becomes more pronounced as the size of the PMDP increases. We also provide comparisons against model-based methods showing that our method, as well as model-free baselines, are more memory-efficient and, therefore, scalable to large MDPs. A preliminary version of this work was presented in \cite{kantaros2022accelerated}. We extend \cite{kantaros2022accelerated} by (i) providing theoretical results that help understand when the proposed approach is, probabilistically, more sample efficient than random exploration methods; 
(ii) providing more comprehensive comparative experiments that do not exist in \cite{kantaros2022accelerated}; and (iii) demonstrating how the biased sampling strategy can be extended to Limit Deterministic Buchi Automata (LDBA) that have smaller state space than DRA and, therefore, can further expedite the learning process \cite{sickert,cai2021optimal,hasanbeig2019reinforcement}. We also release software implementing our proposed algorithm, which can be found in \cite{codeAccRL}.

\textbf{Contribution:} \textit{First}, we propose a novel RL algorithm to quickly learn control policies for \textit{unknown} MDPs with LTL tasks. 
\textit{Second}, we provide conditions under which the proposed algorithm is, probabilistically, more sample-efficient than related works that rely on random exploration. \textit{Third}, we show that the proposed exploration strategy can be employed for various automaton representations of LTL formulas such as DRA and LDBA. \textit{Fourth}, we provide extensive comparative experiments demonstrating the sample efficiency of the proposed method compared to related works. 



%% file: files/PF_short_v3.tex
\subsection{Robot \& Environment Model}
Consider a robot that resides in a partitioned environment with a finite number of states. To capture uncertainty in the robot motion and the workspace, we model the interaction of the robot with the environment as a Markov Decision Process (MDP) of unknown structure, which is defined as follows.
\begin{definition}[MDP]\label{def:labelMDP}
A Markov Decision Process (MDP) is a tuple $\mathfrak{M} = (\ccalX, x_0,  \ccalA, P, \mathcal{AP})$, where $\ccalX$ is a finite set of states; $x_0\in\ccalX$ is an initial state; $\ccalA$ is a finite set of actions. With slight abuse of notation $\ccalA(x)$ denotes the available actions at state $x\in\ccalX$; 
$P:\ccalX\times\ccalA\times\ccalX\rightarrow[0,1]$ is the transition probability function so that $P(x,a,x')$ is the transition probability from state $x\in\ccalX$ to state $x'\in\ccalX$ via control
action $a\in\ccalA$ and $\sum_{x'\in\ccalX}P(x,a,x')=1$, for all $a\in\ccalA(x)$; $\mathcal{AP}$ is a set of atomic propositions; $L : \ccalX \rightarrow 2^{\mathcal{AP}}$ is the labeling function that returns the atomic propositions that are satisfied at a state $x \in \ccalX$. 
\end{definition}

\begin{assumption}[Fully Observable MDP]
We assume that the MDP $\mathfrak{M}$ is fully observable, i.e., at any time step $t$ the current state, denoted by $x_t$, and the observations $L(x_t)\in 2^{\mathcal{AP}}$ in state $x_t$ 
are known.
\end{assumption}

\begin{assumption}[Static Environment]\label{as:label}
\textcolor{black}{We assume that the environment is static in the sense that the atomic propositions that are satisfied at an MDP state $x$ are fixed over time.}
%
\end{assumption}

For instance, Assumption \ref{as:label} implies that obstacles and regions of interest in the environment are static. This assumption can be relaxed using probabilistically labeled MDPs as in \cite{hasanbeig2019reinforcement}.
\vspace{-0.3cm}
\subsection{LTL-encoded Task Specification}
\vspace{-0.1cm}
The robot is responsible for accomplishing a task expressed as an LTL formula, such as sequencing, coverage, surveillance, data gathering or  connectivity tasks \cite{kloetzer2008fully,fainekos2009temporal, leahy2016persistent, guo2017distributed, kantaros2017distributed,fang2022decentralized,vasile2020reactive}. LTL is a formal language that comprises a set of atomic propositions $\mathcal{AP}$, the Boolean operators, i.e., conjunction $\wedge$ and negation $\neg$, and two temporal operators, next $\bigcirc$ and until $\cup$. LTL formulas over a set $\mathcal{AP}$ can be constructed based on the following grammar: $\phi::=\text{true}~|~\pi~|~\phi_1\wedge\phi_2~|~\neg\phi~|~\bigcirc\phi~|~\phi_1~\cup~\phi_2,$ where $\pi\in\mathcal{AP}$. The other Boolean and temporal operators, e.g., \textit{always} $\square$, have their standard syntax and meaning \cite{baier2008principles}. An infinite \textit{word} $w$ over the alphabet $2^{\mathcal{AP}}$ is defined as an infinite sequence  $w=\pi_0\pi_1\pi_2\dots\in (2^{\mathcal{AP}})^{\omega}$, where $\omega$ denotes infinite repetition and $\pi_t\in2^{\mathcal{AP}}$, $\forall t\in\mathbb{N}$. The language $\left\{w\in (2^{\mathcal{AP}})^{\omega}|w\models\phi\right\}$ is defined as the set of words that satisfy the LTL formula $\phi$, where $\models\subseteq (2^{\mathcal{AP}})^{\omega}\times\phi$ is the satisfaction relation \cite{baier2008principles}. In what follows, we consider atomic propositions of the form $\pi^{i}$ that are true if the robot is in state $x_i\in\ccalX$ and false otherwise. 

\vspace{-0.3cm}
\subsection{From LTL formulas to DRA}\label{sec:DRA}
Any LTL formula can be translated into a Deterministic Rabin Automaton (DRA) defined as follows. 

\begin{definition}[DRA \cite{baier2008principles}]\label{def:dra}
A DRA over $2^{\mathcal{AP}}$ is a tuple $\mathfrak{D} = (\ccalQ_D, q_D^0, \Sigma, \delta_D, \ccalF)$, where $\ccalQ_D$ is a finite set of states; $q_D^0 \subseteq Q_D$ is the initial state; $\Sigma=2^{\mathcal{AP}}$ is the input alphabet; $\delta_D: \ccalQ_D \times \Sigma_D \rightarrow {\ccalQ_D}$ is the transition function; and $\ccalF = \{  (\ccalG_1,\ccalB_1), \dots , (\ccalG_f,\ccalB_f) \}$ is a set of accepting pairs where $\ccalG_i, \ccalB_i \subseteq \ccalQ_D, \forall i \in\{1,\dots,f\}$. $\hfill\Box$
\end{definition}
An infinite run $\rho_{D}=q_D^0q_D^1\dots q_D^t\dots$ of $D$ over an infinite word $w=\sigma_0\sigma_1\sigma_2\dots$, where $\sigma_t\in\Sigma$, $\forall t\in\mathbb{N}$, is an infinite sequence of DRA states $q_D^t$, $\forall t\in\mathbb{N}$, such that $\delta(q_D^t,\sigma_t)=q_D^{t+1}$. An infinite run $\rho_D$ is called \textit{accepting} if there exists at least one pair $(\ccalG_i,\ccalB_i)$ such that $\texttt{Inf}(\rho_D)\cap\ccalG_i\neq\varnothing$  and $\texttt{Inf}(\rho_D)\cap\ccalB_i=\varnothing$, where $\texttt{Inf}(\rho_D)$ represents the set of states that appear in $\rho_D$ infinitely often; see also Ex. \ref{ex:dra}.

\begin{figure}
    \centering
    \includegraphics[width=0.7\linewidth]{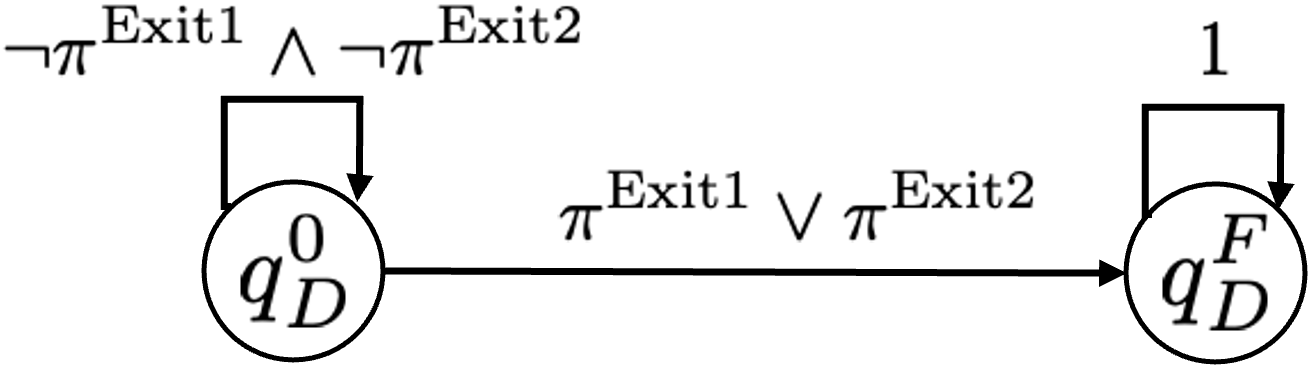}\vspace{-0.3cm}
   \caption{\textcolor{black}{DRA corresponding to $\phi=\Diamond(\pi^{\text{Exit1}}\vee\pi^{\text{Exit2}})$. There is only one set of accepting pairs defined as $\ccalG_1=\{q_D^F\}$ and $\ccalB_1=\{q_D^0\}$. A transition is enabled if the robot generates a symbol satisfying the Boolean formula noted on top of the transitions. All transitions are feasible as per Def. \ref{def:feas}. The function $d_F$ in \eqref{eq:dist2G} is defined as $d_F(q_D^0,\ccalF)=1$ and $d_F(q_D^F,\ccalF)=0$.} }\vspace{-7mm}
    \label{fig:dra}
\end{figure}

\begin{ex}[DRA]\label{ex:dra}
   \textcolor{black}{Consider the LTL formula $\phi=\Diamond(\pi^{\text{Exit1}}\vee\pi^{\text{Exit2}})$ that is true if a robot eventually reaches either Exit1 or Exit2 of a building. The corresponding DRA is illustrated in Figure \ref{fig:dra}.}
\end{ex}

\vspace{-0.3cm}
\subsection{Product MDP}\label{sec:PMDP} 
Given the MDP $\mathfrak{M}$ and the DRA  $\mathfrak{D}$, we define the product MDP (PMDP) $\mathfrak{P}=\mathfrak{M} \times \mathfrak{D}$ as follows.
\begin{definition}[PMDP]\label{def:prodMDP}
Given an MDP $\mathfrak{M} \allowbreak = (\ccalX\allowbreak, x_0\allowbreak,  \ccalA\allowbreak, P\allowbreak, \mathcal{AP}\allowbreak)$ and a DRA $\mathfrak{D} = (\ccalQ_D, q_D^0, \Sigma, \ccalF, \delta_D)$, we define the product MDP (PMDP) $\mathfrak{P}=\mathfrak{M} \times \mathfrak{D}$ as  $\mathfrak{P}\allowbreak = (\mathcal{S}\allowbreak, s_0\allowbreak, \ccalA_\mathfrak{P}\allowbreak, P_{\mathfrak{P}}\allowbreak, \mathcal{F}_\mathfrak{P})$, where 
(i) $\mathcal{S} = \ccalX \times \ccalQ_D$ is the set of states, so that $s=(x,q_D)\in\ccalS$, $x\in\ccalX$, and $q_D\in\ccalQ_D$ ; 
(ii) ${s_0} = (x_0,q_D^0)$ is the initial state; 
(iii) $\mathcal{A}_\mathfrak{P}$ is the set of actions inherited from the MDP, so that $\mathcal{A}_\mathfrak{P} (s) = \ccalA(x)$, where $s=(x,q_D)$;  
(iv) $P_\mathfrak{P}:\ccalS\times\ccalA_\mathfrak{P}\times\ccalS:[0,1]$ is the transition probability function, so that $P_\mathfrak{P}(s,a_P,s')=P(x,a,x'),$
where $s=(x,q_D)\in\ccalS$, $s'=(x',q_D')\in\ccalS$, $a_P\in\ccalA(s)$ and $q_D'=\delta(q,L(x))$; 
(v) $\mathcal{F}_\mathfrak{P} = \{\mathcal{F}_i^\mathfrak{P}\}_{i=1}^{f}$ is the set of accepting states, where $\mathcal{F}_i^\mathfrak{P}$ is a set defined as $\mathcal{F}_i^\mathfrak{P} =  \ccalX \times  \mathcal{F}_i$ and $\mathcal{F}_i=(\ccalG_i,\ccalB_i)$.$\hfill\Box$
\end{definition}

Given  any policy $\boldsymbol\mu:\ccalS\rightarrow \ccalA_\mathfrak{P}$ for $\mathfrak{P}$, we define an infinite run $\rho_{\mathfrak{P}}^{\boldsymbol\mu}$ of $\mathfrak{P}$ to be an infinite sequence of states of $\mathfrak{P}$, i.e., $\rho_{\mathfrak{P}}^{\boldsymbol\mu}=s_0s_1s_2\dots$, where $P_{\mathfrak{P}}(s_t,\boldsymbol\mu(s_t),s_{t+1})>0$. By definition of the accepting condition of the DRA $\mathfrak{D}$, an infinite run  $\rho_{\mathfrak{P}}^{\boldsymbol\mu}$ is accepting 
if there exists at least one pair $i\in\{1,\dots,f\}$ such that $\texttt{Inf}(\rho_{\mathfrak{P}}^{\boldsymbol\mu})\cap\ccalG^\mathfrak{P}_i\neq\emptyset$,
and $\texttt{Inf}(\rho_{\mathfrak{P}}^{\boldsymbol\mu})\cap\ccalB^\mathfrak{P}_i=\emptyset$.

\vspace{-0.3cm}
\subsection{Problem Statement}\label{sec:pr1}
\textcolor{black}{
Our goal is to compute a policy for the PMDP that maximizes the satisfaction probability $\mathbb{P}(\boldsymbol{\mu}\models\phi~|~s_0)$ of an LTL-encoded task $\phi$.
A formal definition of this probability can be found in \cite{baier2008principles,ding2011ltl,guo2018probabilistic}. To this end, 
we first adopt existing reward functions $R:\ccalS\times\ccalA_{\mathfrak{P}}\times\ccalS\rightarrow\mathbb{R}$ defined based on the accepting condition of the PMDP as e.g., in \cite{gao2019reduced}. 
Then, our goal is to compute a policy $\boldsymbol\mu^*$ that maximizes the expected accumulated return, i.e., $\boldsymbol\mu^*(s)=\arg\max\limits_{\boldsymbol\mu \in \mathcal{D}}~ {U}^{\boldsymbol\mu}(s)$,
where $\mathcal{D}$ is the set of all stationary deterministic policies over $\mathcal{S}$, and 	
\begin{equation}\label{eq:utility}
{U}^{\boldsymbol\mu}(s)=\mathbb{E}^{\boldsymbol\mu} [\sum\limits_{t=0}^{\infty} \gamma^t~ R(s_t,\boldsymbol\mu(s_t),s_{t+1})|s=s_0].
\end{equation}
In \eqref{eq:utility}, $\mathbb{E}^{\boldsymbol\mu} [\cdot]$ denotes the expected value given that the  PMDP follows the policy $\boldsymbol\mu$ \cite{puterman}, $0\leq\gamma< 1$ is the discount factor, and $s_0,...,s_t$ is the sequence of states generated by $\boldsymbol\mu$ \textcolor{black}{up to time $t$}, initialized at $s_0$. Since the PMDP has a finite state/action space and $\gamma<1$, there exists a stationary deterministic optimal policy $\boldsymbol\mu^*$ \cite{puterman}.  
The reward function $R$ and the discount factor $\gamma$ should be designed so that maximization of \eqref{eq:utility} is equivalent to maximization of the satisfaction probability. Efforts towards this direction are presented e.g., in \cite{gao2019reduced,hasanbeig2019reinforcement} while provably correct rewards and discount factors for PMDPs constructed using LDBA, instead of DRA, are proposed in \cite{bozkurt2020control,hasanbeig2023certified,xuan2024uniqueness}. However, as discussed in Section \ref{sec:Intro}, due to sparsity of these rewards, these methods are sample-inefficient. This is the main challenge that this paper aims to address.} 
\begin{problem}\label{pr:pr1}
Given (i) an MDP $\mathfrak{M}$ with unknown transition probabilities and underlying graph structure; (ii) a task specification captured by an LTL formula $\phi$; (iii) a reward function $R$ for the PMDP motivating satisfaction of its accepting condition, develop a sample-efficient RL algorithm that can learn a deterministic control policy $\boldsymbol\mu^*$ that maximizes \eqref{eq:utility}.
\end{problem}


%% file: files/alg2.tex





\normalsize

To solve Problem \ref{pr:pr1}, we propose a new reinforcement learning (RL) algorithm that can quickly synthesize control policies that maximize \eqref{eq:utility}. The proposed algorithm is summarized in Algorithm \ref{alg:RL-LTL} and described in detail in the following subsections. First, in Section \ref{sec:prune}, we define a distance function over the DRA state-space. 
%
In Sections \ref{sec:accRL}--\ref{sec:biasedExpl}, we describe the proposed logically-guided RL algorithm for LTL control objectives. To accelerate the learning phase, the distance function defined in Section \ref{sec:prune} is utilized to guide exploration. A discussion on how the proposed algorithm can be applied to LDBA, that typically have a smaller state space than DRA, is provided in Appendix \ref{app:ldba}.


\vspace{-0.3cm}
\subsection{Distance Function over the DRA State Space}\label{sec:prune}
First, the LTL task $\phi$ is converted into a DRA; see Definition \ref{def:dra} [line \ref{rl:dra}, Alg. \ref{alg:RL-LTL}].
Then, we define a distance-like function over the DRA state-space that measures how 'far' the robot is from accomplishing the assigned LTL tasks [line \ref{rl:dist}, Alg. \ref{alg:RL-LTL}]. In other words, this function returns how far any given DRA state is from the sets of accepting states $\ccalG_i$. To define this function, first, we remove from the DRA all infeasible transitions that cannot be physically enabled. To define infeasible transitions, we first define feasible symbols as follows \cite{kantaros2020stylus}\textcolor{black}{; see Fig. \ref{fig:dra}.} 
%
\begin{definition}[Feasible symbols $\sigma\in\Sigma$]\label{def:feas}
A symbol $\sigma\in\Sigma$ is \textit{feasible} if and only if $\sigma\not\models b^{\text{inf}}$, where $b^{\text{inf}}$ is a Boolean formula defined as $b^{\text{inf}}=\vee_{\forall x_i\in\ccalX}( \vee_{\forall x_e\in\ccalX\setminus\{x_i\}}(\pi^{x_i}\wedge\pi^{x_e}))$,
where $b^{\text{inf}}$ requires the robot to be present in more than one MDP state simultaneously. All feasible symbols $\sigma$ are collected in a set denoted by $\Sigma_{\text{feas}}\subseteq\Sigma$.$\hfill\Box$
\end{definition}

Then, we prune the DRA by removing infeasible DRA transitions defined as follows:
\begin{definition}[Feasibility of DRA transitions]
A DRA transition from $q_D$ to $q_D'$ is feasible if there exists at least one feasible symbol $\sigma\in\Sigma_{\text{feas}}$ such that $\delta(q_D,\sigma)=q_D'$; otherwise, it is infeasible.$\hfill\Box$
\end{definition}
Next,  we define a function $d:\ccalQ_D\times\ccalQ_D\rightarrow \mathbb{N}$ that returns the minimum number of \textit{feasible} DRA transitions required to reach a state $q_D'\in\ccalQ_D$ starting from a state $q_D\in\ccalQ_D$. 
Particularly, we define this function 
as follows \cite{kantaros2020stylus,ding2014ltl}:
\begin{equation}\label{eq:dist1}
d(q_D,q_D')=\left\{
                \begin{array}{ll}
                  |SP_{q_D,q_D'}|, \mbox{if $SP_{q_D,q_D'}$ exists,}\\
                  \infty, ~~~~~~~~~\mbox{otherwise},
                \end{array}
              \right.
\end{equation}
where $SP_{q_D,q_D'}$ denotes the shortest path (in terms of hops) in the  pruned DRA from $q_D$ to $q_D'$ and $|SP_{q_D,q_D'}|$ stands for its cost (number of hops). Note that if $d(q_D^0,q_D)= \infty$, for all $q_D\in\ccalG_i$ and for all $i\in\{1,\dots,n\}$, then the LTL formula can not be satisfied since the in the pruning process, only the DRA transitions that are impossible to enable are removed. 
Next, using \eqref{eq:dist1}, we define the following distance function:\footnote{Observe that, unlike \cite{lacerda2015optimal,ding2014optimal}, $ d_F(q_D,\ccalF)$ may not be equal to $0$ even if $q_D\in\ccalG_i$. The latter may happen if $q_D$ does not have a feasible self-loop.}
\begin{equation}\label{eq:dist2G}
    d_F(q_D,\ccalF)=\min_{q_D^G\in\cup_{i\in\{1,\dots,f\}}\ccalG_i} d(q_D,q_D^G).
\end{equation}
In words, \eqref{eq:dist2G} measures the distance from any DRA state $q_D$ to the set of accepting pairs, i.e., the distance to the closest DRA state $q_D^G$ that belongs to $\cup_{i\in\{1,\dots,f\}}\ccalG_i$; \textcolor{black}{see also Fig. \ref{fig:dra}.}

\begin{algorithm}[t]
\caption{Accelerated RL for LTL Control Objectives}
\label{alg:RL-LTL}
\begin{algorithmic}[1]
\State Initialize: (i) $Q^{\boldsymbol\mu}(s,a)$ arbitrarily, (ii) $\hat{P}(x,a,x')=0$, (iii) $c(x,a,x')=0$, (iv) $n(x,a)=0$, for all $x,x'\in\ccalX$ and $a\in\ccalA(x)$, and  (v) \textcolor{black}{$n_{\mathfrak{P}}(s,a,s')=0$} for all $s,s'\in\ccalS$ and $a\in\ccalA_{\mathfrak{P}}(s)$;\label{rl:intit0}
\State Convert $\phi$ to a DRA $\mathcal{D}$;\label{rl:dra}
\State Construct distance function $d_F$ over the DRA as per \eqref{eq:dist2G}; \label{rl:dist}
\State $\boldsymbol\mu=(\epsilon,\delta)-\text{greedy}(Q_{})$;\label{rl:init1}
\State $\texttt{episode-number}=0$; \label{rl:init3}
\While{$Q$ has not converged}\label{rl:while1}
\State $\texttt{episode-number} = \texttt{episode-number} + 1$; \label{rl:epi}
\State	Initialize \textcolor{black}{time step} $t=0$;
\State  Initialize $s_t=(x_0,q_D^0)$ for a randomly selected $x_0$; \label{rl:initState} 
 \label{rl:initIterEpi}
        \While{$t<\tau$} \label{rl:while2}
        \State Pick action $a_t$ as per \eqref{eq:policy};\label{rl:pickAction}
        \State Execute $a_{t}$ and observe ${s}_{t+1}=(x_{t+1},q_{t+1})$, and $R({s}_t,a_t,{s}_{t+1})$; \label{rl:exec}
        \State $n({x}_t,a_t) = n({x}_t,a_t) + 1$;\label{rl:incrCounterStAct}
        \State $c({x}_t,a_t,{x}_{t+1}) = c({x}_t,a_{t},{x}_{t+1}) + 1$; \label{rl:incrCounterStActSt}
        \State Update $\hat{P}({x}_t,a_{t},{x}_{t+1})$ as per \eqref{eq:probEst};\label{rl:probEst}
        \State $n_{\mathfrak{P}}({s}_t,a_{t}) = n_{\mathfrak{P}}({s}_t,a_{t}) + 1$; \label{rl:incrCounterTrans}
        \State Update $Q^{\boldsymbol\mu}({s}_t,a_{t})$ as per \eqref{eq:Qupd};\label{rl:updQ}
        \State ${s}_t={s}_{\text{next}}$; \label{rl:resetSt}
        \State $t= t + 1$; \label{rl:iter0}
    	\State Update $\epsilon, \delta_b, \delta_e$;  \label{rl:iter}
    	\EndWhile
\EndWhile
\end{algorithmic}
\end{algorithm}

\vspace{-0.3cm}
\subsection{Learning Optimal Temporal Logic Control Policies}\label{sec:accRL}
\vspace{-0.1cm}

In this section, we present the proposed accelerated RL algorithm for LTL control synthesis [lines \ref{rl:init1}-\ref{rl:iter}, Alg. \ref{alg:RL-LTL}]. The output of the proposed algorithm is a \textit{stationary deterministic} policy $\boldsymbol\mu^*$ 
for $\mathfrak{P}$ maximizing \eqref{eq:utility}. 
To construct $\boldsymbol\mu^*$, we employ episodic Q-learning (QL). 
%
%
Similar to standard QL, starting from an initial PMDP state, we define learning episodes over which the robot picks actions as per a stationary and stochastic control policy $\boldsymbol\mu:\ccalS\times \ccalA_{\mathfrak{P}}\rightarrow [0,1]$ that eventually converges to $\boldsymbol\mu^*$ [lines \ref{rl:init1}-\ref{rl:init3}, Alg. \ref{alg:RL-LTL}]. Each episode terminates after a user-specified number of \textcolor{black}{time steps} $\tau$ or if the robot reaches a deadlock PMDP state, i.e., a state with no outgoing transitions [lines \ref{rl:epi}-\ref{rl:iter}, Alg. \ref{alg:RL-LTL}]. \textcolor{black}{Notice that the hyper-parameter $\tau$ should be selected to be large enough to ensure that the agent learns how to repetitively visit the accepting states \cite{bozkurt2020control,hasanbeig2019reinforcement,hasanbeig2022lcrl}.}
The RL algorithm terminates once an action value function $Q^{\boldsymbol\mu}(s,a)$ has converged. This action value function is defined as the expected return for taking
action $a$ when at state $s$ and then following policy $\boldsymbol\mu$ \cite{rlbook}, i.e., 
\begin{equation}\label{eq:Q}
Q^{\boldsymbol\mu}(s,a)=\mathbb{E}^{\boldsymbol\mu} [\sum\limits_{t=0}^{\infty} \gamma^t~ R(s_t,\boldsymbol\mu(s_t),s_{t+1})|s_0=s, a_0=a].
\end{equation}
We have that  $U^{\boldsymbol\mu}(s)=\max_{a \in \mathcal{A}_\mathfrak{P}(s)}Q^{\boldsymbol\mu}(s,a)$ \cite{rlbook}. The action-value function $Q^{\boldsymbol\mu}(s,a)$ can be initialized arbitrarily. 

During any learning episode the following process is repeated until the episode terminates. First, given the  PMDP state $s_{t}$ at the \textcolor{black}{current time step $t$}, initialized as $s_{t}=s_0$ [line \ref{rl:initState}, Alg. \ref{alg:RL-LTL}], an action $a_{t}$ is selected as per a policy $\boldsymbol\mu$ [line \ref{rl:pickAction}, Alg. \ref{alg:RL-LTL}]; the detailed definition of $\boldsymbol\mu$ will be given later. The selected action is executed yielding the next state ${s}_{t+1}=(x_{t+1},q_{t+1})$, and a reward $R({s}_{t},a_{t},{s}_{t+1})$.
For instance, the reward function $R$ can be constructed as in \cite{gao2019reduced} defined as follows:
\begin{equation}\label{eq:RewardQ}
R(s,a_\mathfrak{P},s')=\left\{
                \begin{array}{ll}
                  r_{\ccalG}, ~\mbox{if $s'\in\ccalG_i^{\mathfrak{P}}$,}\\
                  r_{\ccalB}, ~\mbox{if $s'\in\ccalB_i^{\mathfrak{P}}$,}\\
                  r_d, ~\mbox{if $s'$ is a deadlock state,}\\
                  r_0, ~\mbox{otherwise,}
                \end{array}
              \right.
\end{equation}
In \eqref{eq:RewardQ}, we have that $r_{\ccalG}>0$, for all $i\in\{1,\dots,f\}$, and $r_d<r_{\ccalB}<r_0\leq0$. This reward function motivates the robot to satisfy the PMDP accepting condition, i.e., to visit the states $\ccalG_j^{\mathfrak{P}}$ as often as possible and minimize the number of times it visits $\ccalB_i^{\mathfrak{P}}$ and deadlock states while following the shortest possible path; deadlock states are visited when the LTL task is violated, e.g., when collision with an obstacle occurs.


Given the new state ${s}_{t+1}$, the MDP model of the robot is updated. In particular, every time an MDP transition is enabled, the corresponding transition probability is updated. Let $\hat{P}(x_t,a_t,x_{t+1})$ denote the estimated MDP transition probability from state $x_t\in\ccalX$ to state $x_{t+1}\in\ccalX$, when an action $a$ is taken. These estimated MDP transition probabilities are initialized so that  $\hat{P}(x,a,x')=0$, for all combinations of states and actions, and they are continuously updated at \textcolor{black}{every time step $t$} of each episode as [lines \ref{rl:incrCounterStAct}-\ref{rl:probEst}]:
\begin{equation}\label{eq:probEst}
    \hat{P}(x_{t},a_{t},x_{t+1})=\frac{c(x_{t},a_{t},x_{t+1})}{n(x_{t},a_{t})},
\end{equation}
where (i) $n:\mathcal{X}\times\mathcal{A}\rightarrow\mathbb{N}$ is a function that returns the number of times action $a$ has been taken at an MDP state $x$ and (ii) $c :\mathcal{X}\times\mathcal{A}\times\ccalX\rightarrow\mathbb{N}$ is a function that returns the number of times an MDP state $x'$ has been visited after taking action $a$ at a state $x$. Note that as $n(x,a)\to\infty$ the estimated transition probabilities $\hat{P}(x,a,x')$ converge asymptotically to the true transition probabilities $P(x,a,x')$, for all transitions. 

Next, the action value function is updated as follows \cite{rlbook} [line \ref{rl:updQ}, Alg. \ref{alg:RL-LTL}] :
\begin{align}\label{eq:Qupd}
    &Q^{\boldsymbol\mu}({s}_t,a_{t})= Q^{\boldsymbol\mu}({s}_t,a_{t})+(1/n_{\mathfrak{P}}({s}_t,a_{t}))&\nonumber\\&[R({s}_t,a_{t})-Q^{\boldsymbol\mu}({s}_t,a_{t})+\gamma \max_{a'}Q^{\boldsymbol\mu}({s}_{t+1},a'))],
\end{align}
where $n_{\mathfrak{P}}:\mathcal{S}\times\mathcal{A}_\mathfrak{P}\rightarrow\mathbb{N}$ counts the number of times that action $a$ has been taken at the PMDP state $s$. Once the action value function is updated, the current PMDP state is updated as $s_{t}=s_{t+1}$, \textcolor{black}{the time step $t$} is increased by one, and the policy $\boldsymbol\mu$ gets updated [lines \ref{rl:resetSt}-\ref{rl:iter}, Alg. \ref{alg:RL-LTL}]. 


%
%
%

As a policy $\boldsymbol\mu$, we propose an extension of the $\epsilon$-greedy policy, called $(\epsilon,\delta)$-greedy policy, that selects an action $a$ at an PMDP state $s$ by using the learned action-value function $Q^{\boldsymbol\mu}(s,a)$ and the continuously learned transition probabilities $\hat{P}(x,a,x')$. Formally, the $(\epsilon,\delta)$-greedy policy $\boldsymbol\mu$ is defined as
%
\begin{equation}\label{eq:policy}
\boldsymbol\mu(s,a) = \begin{cases}
1-\epsilon + \frac{\delta_e}{|\mathcal{A}_\mathfrak{P}(s)|} &\text{if~}  \textcolor{black}{a=a^*~\text{and~} a\neq a_b}, \\
1-\epsilon + \frac{\delta_e}{|\mathcal{A}_\mathfrak{P}(s)|} +\delta_b &\text{if~}  \textcolor{black}{a=a^*~\text{and~} a=a_b,} \\
\delta_e/|\mathcal{A}_\mathfrak{P}(s)|  &\text{if}~\textcolor{black}{a \in \mathcal{A}_\mathfrak{P}(s)\setminus\{a^*,a_b\}},\\
\delta_b + \delta_e/|\mathcal{A}_\mathfrak{P}(s)| &\text{if~} \textcolor{black}{a=a_b \text{~and~} a\neq a^*,}
\end{cases}
\end{equation}
where \textcolor{black}{$\delta_b,\delta_e\in[0,1]$ and $\epsilon = \delta_b+\delta_e\in[0,1]$}. In words, according to this policy, (i) with probability $1-\epsilon$, the \textit{greedy} action $a^*=\argmax_{a\in\mathcal{A}_\mathfrak{P}}Q(s,a)$ is taken (as in the standard $\epsilon$-greedy policy); and (ii) an exploratory action is selected with probability $\epsilon=\delta_b+\delta_e$. The exploration strategy is defined as follows: (ii.1) with probability $\delta_e$ a random action $a$ is selected (\textit{random} exploration); and (ii.2) with probability $\delta_b$ the action, denoted by $a_b$, that is most likely to drive the robot towards an accepting product state in $\ccalG_i^{\mathfrak{P}}$ is taken (\textit{biased} exploration). The action $a_b$ will be defined formally in Section \ref{sec:biasedExpl}.
As in standard QL, $\epsilon$ should asymptotically converge to $0$ while ensuring that eventually all actions have been applied infinitely often at all states. This ensures that $\boldsymbol\mu$ asymptotically converges to the optimal greedy policy
\begin{equation}\label{eq:greedy}
\boldsymbol\mu^*=\argmax_{a\in\mathcal{A}_\mathfrak{P}} Q^*(s,a)   
\end{equation}
where $Q^*$ is the optimal action value function; see  Sec. \ref{sec:polImprov}. We note that $Q^{\boldsymbol\mu^*}(s,\boldsymbol\mu^*(s))=U^{\boldsymbol\mu^*}(s)=V^*({s})$, where $V^*({s})$ is the optimal value function that could have been computed 
if the MDP was fully known \cite{rlbook,reachability_in_hybrid}.
Given $\epsilon$, selection of the parameters $\delta_e$ and $\delta_b$ is discussed in Sec. \ref{sec:theory}. 
%
%
\vspace{-0.4cm}
\subsection{Specification-guided Exploration for Accelerated Learning}\label{sec:biasedExpl}
\begin{figure}
    \centering
     \includegraphics[width=0.8\linewidth]{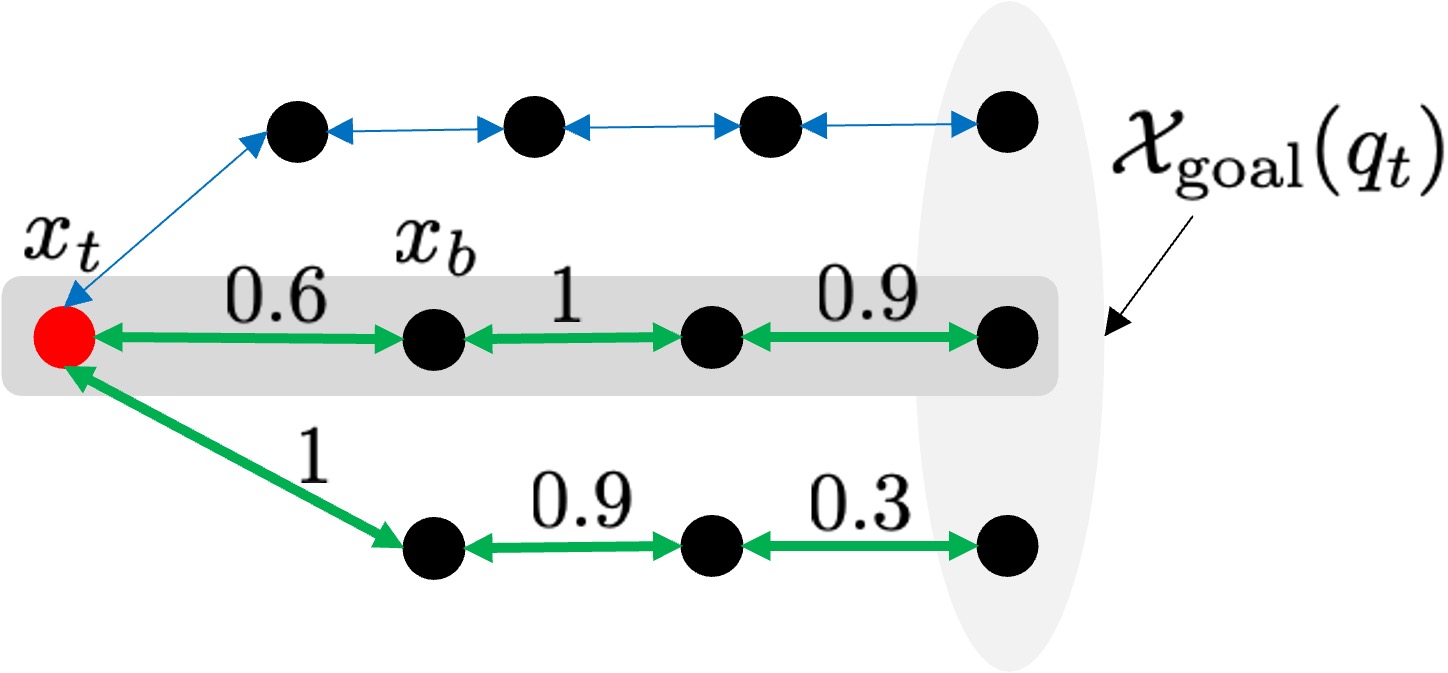}\vspace{-0.3cm}
    \caption{\textcolor{black}{Graphical depiction of the sets $\ccalX_{\text{goal}}(q_{t})$. The disks represent MDP states and the arrows between states mean that there exists at least one action such that the transition probability from one state to another one is non-zero. 
    The length of the shortest path from $x_t$ to $\ccalX_{\text{goal}}$ is $3$ hops, i.e., $J_{x_t,\ccalX_{\text{goal}}}=3$; see \eqref{eq:dist2set}. Also, the paths $p_j^t$, $j\in\ccalJ=\{1,2\}$ are highlighted with thick green lines. The numbers on top of the green edges represent $\max_{a}P(p_j^t(e),a,p_j^t(e+1))$; see \eqref{eq:optUnc}. Observe that $p^*$ is the green path highlighted with gray color. }
    }\vspace{-5mm}
    \label{fig:notations1}
\end{figure}

Next, we describe the design of the biased action $a_b$ in \eqref{eq:policy}. First, we need to introduce the following definitions; see Fig. \ref{fig:notations1}. Let $s_{t}=(x_{t},q_{t})$ denote the current PMDP state at the current learning episode and \textcolor{black}{time step $t$} of Algorithm \ref{alg:RL-LTL}. 
%
Let $\ccalQ_{\text{goal}}(q_{t})\subset\ccalQ$ be a set that collects all DRA states that are one-hop reachable from $q_{t}$ in the pruned DRA and they are closer to the accepting DRA states than $q_{t}$ is, as per \eqref{eq:dist2G}. \textcolor{black}{Formally}, $\ccalQ_{\text{goal}}(q_{t})$ is defined as follows:
\begin{align}\label{eq:Qnext}
    &\ccalQ_{\text{goal}}(q_{t})=\{q'\in\ccalQ~|~(\exists\sigma\in\Sigma_{\text{feas}}~\text{such that}\\&\delta_D(q_{t},\sigma)=q') \wedge (d_F(q',\ccalF)=d_F(q_{t},\ccalF)-1)\}.\nonumber
\end{align}
\vspace{-0.5cm}

Also, let $\ccalX_{\text{goal}}(q_{t})\subseteq\ccalX$ be a set of MDP states, denoted by $x_{\text{goal}}$, that if the robot eventually reaches, then transition from $s_{t}$ to a product state $s_{\text{goal}}=[x_{\text{goal}},q_{\text{goal}}]$ will occur, where  $q_{\text{goal}}\in\ccalQ_{\text{goal}}(q_{t})$; \textcolor{black}{see also Ex. \ref{ex:pjk}.} Formally, $\ccalX_{\text{goal}}(q_{t})$ is defined as follows:
\begin{align}\label{eq:Xgoal}
\ccalX_{\text{goal}}(q_{t}) = \{&x\in\ccalX~|~\delta_D(q_{t},L(x))\in\ccalQ_{\text{goal}}(q_{t})\}.
\end{align}
Next, we view the continuously learned MDP as a weighted directed graph $\ccalG=(\ccalV,\ccalE, w)$ where the set $\ccalV$ is the set of MDP states, $\ccalE$ is the set of edges, and $w:\ccalE\rightarrow \mathbb{R}_{+}$ is function assigning weights to each edge. Specifically, an edge from the node (MDP state) $x$ to $x'$ exists if there exists at least one action $a\in\ccalA(x)$ such that $\hat{P}(x,a,x')>0$. Hereafter, we assign a weight equal to $1$ to each edge; see also Remarks \ref{rem:shortestPath}-\ref{rem:weight}. 
\textcolor{black}{We denote the cost of the shortest path from $x$ to $x'$ by $J_{x,x'}$.  Next, we define the cost of the shortest path connecting a state $x$ to the set $\ccalX_{\text{goal}}$ as follows:
\begin{equation}\label{eq:dist2set} J_{x,\ccalX_{\text{goal}}}=\min_{x'\in\ccalX_{\text{goal}}} J_{x,x'}.
\end{equation}
Let $J$ be the total number of paths from $x$ to $\ccalX_{\text{goal}}$, where their length (i.e., number of hops) is $J_{x,\ccalX_{\text{goal}}}$. 
%
%
We denote such a path by $p_j^t$, $j\in\ccalJ:=\{1,\dots,J\}$, and the $e$-th MDP state in this path by $p_j^t(e)$. Then, among all the paths $p_j^t$, we compute the one with the minimum uncertainty-based cost $C(p_j^t)$; see Fig. \ref{fig:notations1}. We define this cost as 
%
%
\begin{equation}\label{eq:uncCost}
  C(p_j^t)=\prod_{e=1}^{J_{x,\ccalX_{\text{goal}}}} \left[\max_{a}\hat{P}(p_j^t(e),a,p_j^t(e+1))\right],
\end{equation}
where the maximization is over all actions $a\in\ccalA(p_j^t(e))$. We denote by $p^*$ the path with the minimum cost $C(p_j^t)$, i.e., $p^*=p_{j^*}^t$, where $j^*=\argmax_j  C(p_j^t)$. Thus, we have that:
\begin{equation}\label{eq:optUnc}
    C(p^*)\geq C(p_j^t), \forall j\in\ccalJ.
\end{equation}
}
\textcolor{black}{Once $p^*$ is constructed, the action $a_b$ is defined as follows:
\begin{equation}\label{eq:a_b}
 a_b = \argmax_{a\in \ccalA(x_{t})}\ \hat{P}(x_{t},a,x_b),
\end{equation} 
where $x_b=p^*(2)$; see Fig. \ref{fig:notations1}. In words, $a_b$ is the action with the highest probability of allowing the system to reach the state $p^*(2)$, i.e., to move along the best path $p^*$. Observe that computation of the biased action does not depend on the employed reward structure nor on perfectly learning all MDP transition probabilities.}

\begin{rem}[Initialization]
\textcolor{black}{Selection of the biased action $a_b$  requires knowledge of (i) the MDP states $x$ in \eqref{eq:Xgoal} that need to be visited to enable transitions to DRA states in $\ccalQ_{\text{goal}}$; and (ii) the underlying MDP graph structure, determined by the (unknown) transition probabilities, to compute \eqref{eq:dist2set}. However, neither of them may be available in early episodes. In this case, we randomly initialize $\ccalX_{\text{goal}}$ for (i). Similarly, for (ii), the estimated transition probabilities are randomly initialized (or, simply, set equal to $0$ [line \ref{rl:intit0}, Alg. \ref{alg:RL-LTL}]) initializing this way the MDP graph structure. If no paths can be computed to determine $J_{x_t,\ccalX_{\text{goal}}}$ in \eqref{eq:dist2set}, we select a random biased action.}
\end{rem}
\begin{rem}[Computing Shortest Path]\label{rem:shortestPath}
It is possible that the shortest path from $x_{t}$ to $x_{\text{goal}}\in\ccalX_{\text{goal}}(q_{t})$ goes through states/nodes $x$ that if visited, a transition to a new state $q\neq q_{t}$ that does not belong to $\ccalQ_{\text{goal}}(q_{t})$ may be enabled. Therefore, when we compute the shortest paths, 
we treat all such nodes $x$ as `obstacles' that should not be crossed. These states are collected in the set $\ccalX_{\text{avoid}}$ defined as $\ccalX_{\text{avoid}}=\{x\in\ccalX~|~\delta(q_{t},L(x))=q_D\notin\ccalQ_{\text{goal}}\}$.
\end{rem}
\begin{rem}[Weights \& Shortest Paths]\label{rem:weight}
\textcolor{black}{To design the biased action $a_b$, the MDP is viewed as a weighted graph where a weight $w=1$ is assigned to all edges. 
%
In Section \ref{sec:theory}, this definition of weights allows us to show how the probability of making progress towards satisfying the assigned task (i.e., reaching the DRA states $\ccalQ_{\text{goal}}$) within the minimum number of time steps (i.e., $J_{x_t,\ccalX_{\text{goal}}}$ time steps) is positively affected by introducing bias in the exploration phase.
Alternative weight assignments can be used that may further improve sample-efficiency in practice; see also Ex. \ref{ex:pjk}. For instance, the assigned weights can be equal to the reciprocal of the estimated transition probabilities. In this case, the shortest path between two MDP states models the path with the least uncertainty that connects these two states. However, in this case the theoretical results shown in Section \ref{sec:theory} do not hold.
}
\end{rem}

\begin{figure}
    \centering
    \includegraphics[width=1\linewidth]{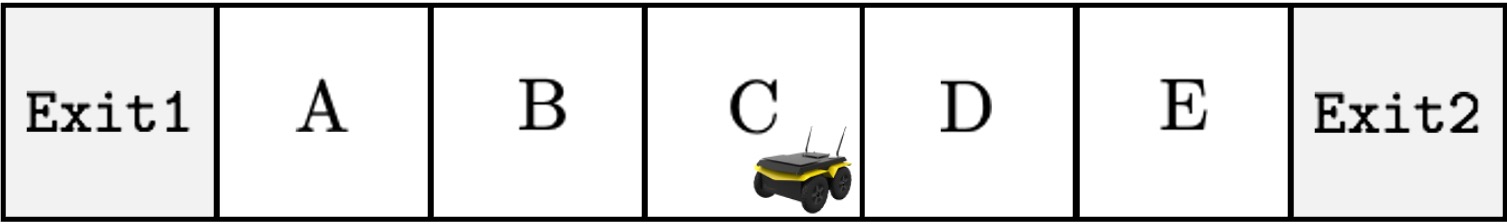}\vspace{-0.3cm}
   \caption{MDP-based representation of the interaction of a ground robot with corridor-like environment. The square cells represent MDP states, i.e., $\ccalX=\{\text{Exit1},\text{Exit2}, \text{A}, \text{B},\text{C}, \text{D}, \text{E}\}$. An action enabling transition between adjacent cells with non-zero probability exists for all MDP states.}\vspace{-0.7cm}
    \label{fig:example1}
\end{figure}

\begin{ex}[Biased Exploration]\label{ex:pjk}
\textcolor{black}{
Consider a robot operating in a corridor of a building as in Figure \ref{fig:example1}.  The robot is tasked with exiting the building i.e., eventually reaching one of the two exits. This can be captured by the following LTL formula: $\phi=\Diamond(\pi^{\text{Exit1}}\vee\pi^{\text{Exit2}})$. The DRA of this specification is illustrated in Figure \ref{fig:dra}. Assume that $q_t=q_D^0$. Then, $\ccalX_{\text{goal}}=\{\text{Exit1}, \text{Exit2}\}$.
The robot can take two actions at each state (besides the `exit' states): $a_1=\text{`left'}$ and $a_2=\text{`right'}$.  \textbf{(i)} Assume that $x_t=C$. Observe that   $J_{x_t,\ccalX_{\text{goal}}}=3$ and that $J=2$. Specifically, the following two paths $p_j^t$ can be defined: $p_1^t=\text{C,D,E,Exit1}$ and $p_2^t=\text{C,B,A,Exit2}$. Consider also  transition probabilities that satisfy $\max_a P(C,a,D)=0.51$, $\max_a P(D,a,E)=0.9$, $\max_a P(E,a,\text{Exit2})=1$, $\max_a P(C,a,B)=0.9$, $\max_a P(B,a,A)=0.6$, $\max_a P(A,a,\text{Exit1})=0.6$. In this case, we have that $C(p_1^t)=0.459$ and $C(p_2^t)=0.324$. According to \eqref{eq:optUnc}, we have that $j^*=1$ and, therefore, $x_b=p_1^t(2)=\text{D}$. The biased action $a_b$ at $x_t$ is  $a_b=a_2$ as per \eqref{eq:a_b}. 
\textbf{(ii)} Assume that $x_t=D$. Then, we have that $J_{x_t,\ccalX_{\text{goal}}}=2$. Notice that there is only path to reach $\ccalX_{\text{goal}}$ within $J_{x_t,\ccalX_{\text{goal}}}=2$ hops/time steps defined as $p_1^t=\text{D,E,Exit1}$. Consider also transition probabilities that satisfy  $\max_a P(D,a,E)=0.7$, $\max_a P(E,a,\text{Exit2})=0.7$, $\max_a P(D,a,C)=1$, $\max_a P(C,a,B)=1$, $\max_a P(B,a,A)=1$, $\max_a P(A,a,\text{Exit1})=1$. In this case, we have that $C(p_1^t)=0.49$. The biased action $a_b$ at $x_t$ is selected as follows. Assume $P(D,a_1,E)=0.3$ and $P(D,a_2,E)=0.7$. Given that $x_b=p_1^t(2)=\text{E}$, we have that $a_b=a_2$ as per \eqref{eq:a_b}. Observe that although there is a `deterministic' path from $x_t$ to \text{Exit1} of length $4$ that can be followed with probability $1$, the biased action aims to drive the robot towards Exit2. This happens because the proposed algorithm is biased towards the shortest paths (of length $2$ here), in terms of number of MDP transitions/hops, that will lead to DRA states that are closer to the accepting states by definition of the weights $w$. We note that the paths stemming from the biased action are not necessarily the paths with the least uncertainty; see also Rem. \ref{rem:weight}. Also, we highlight that we do not claim any optimality of $a_b$ with respect to the task satisfaction probability; intuitively, in (ii), the biased action is `sub-optimal' with respect to the task satisfaction probability.}
 \end{ex}

%% file: files/theory.tex


In this section, we show that any $(\epsilon,\delta)-$greedy policy achieves policy improvement; see Proposition \ref{prop:polImprov}. Also, we provide conditions that $\delta_b$ and $\delta_e$ should satisfy under which the proposed biased exploration strategy results in learning control policies faster, in a probabilistic sense, than policies that rely on uniform-based exploration. We emphasize that these results should be interpreted primarily in an existential way as they rely on the unknown MDP transition probabilities. 
First, we provide `myopic' sample-efficiency guarantees. Specifically, we show that starting from $s_t=(x_t,q_t)$, the probability of reaching PMDP states $s_{t+1}=(x_{t+1},q_{t+1}))$, where $x_{t+1}$ is closer to $\ccalX_{\text{goal}}$ (see \eqref{eq:Xgoal}) than $x_t$, is higher when bias is introduced in the exploration phase; see Section \ref{sec:myopic}. Then, we provide non-myopic guarantees that ensure that starting from $s_t$ the probability of reaching PMDP states $s_{t'}=(x_{t'},q_{t'})$, where $t'>t$ and $q_{t'}\in\ccalQ_{\text{goal}}$ (see \eqref{eq:Qnext}), in the minimum number of time steps (as determined by $J_{x_t,\ccalX_{\text{goal}}}$) is higher when bias is introduced in the exploration phase; see Section \ref{sec:nonmyopicBias}. 
%

\vspace{-0.3cm}
\subsection{Policy Improvement}\label{sec:polImprov}
\begin{prop}[Policy Improvement]\label{prop:polImprov}
For any $(\epsilon,\delta)$-greedy policy $\boldsymbol{\mu}$, the updated $(\epsilon,\delta)$-greedy policy $\boldsymbol{\mu'}$ obtained after updating the state-action value function $Q^{\boldsymbol{\mu}}(s,a)$ satisfies $U^{\boldsymbol{\mu'}}(s)\geq U^{\boldsymbol{\mu}}(s)$, for all $s\in\ccalS$.
$\hfill\Box$
\end{prop}
\vspace{0cm}
\begin{proof}
To show this result, we follow the same steps as in the policy improvement result for the $\epsilon$-greedy policy \cite{rlbook}. For simplicity of notation, hereafter we use $A=|\ccalA_{\mathfrak{P}}(s)|$. Thus, we have that:
$ U^{\boldsymbol\mu'}(s)=\sum_{a\in\ccalA_{\mathfrak{P}}(s)}\boldsymbol\mu'(s,a)Q^{\boldsymbol\mu}(s,a)=\frac{\delta_{e}}{A}\sum Q^{\boldsymbol\mu}(s,a)+(1-\epsilon)\max_{a\in\ccalA_{\mathfrak{P}}(s)}Q^{\boldsymbol\mu}(s,a)+\delta_b Q^{\boldsymbol\mu}(s,a_b)\geq\frac{\delta_{e}}{A}\sum Q^{\boldsymbol\mu}(s,a)+(1-\epsilon)\sum_{a\in\ccalA_{\mathfrak{P}}(s)}\left(\frac{\boldsymbol\mu(s,a)-\frac{\delta_e}{A}-\mathbb{I}_{a=a_b}\delta_b}{1-\epsilon}\right)Q^{\boldsymbol\mu}(s,a)+\delta_b Q^{\boldsymbol\mu}(s,a_b)=\sum_{a\in\ccalA_{\mathfrak{P}}(s)}\boldsymbol\mu(s,a)Q^{\boldsymbol\mu}(s,a)=U^{\boldsymbol\mu}(s)$
where the inequality holds because the summation is a weighted average with non-negative weights  summing to $1$, and as such it must be less than the largest number averaged.
\end{proof}

In Proposition \ref{prop:polImprov}, the equality $U^{\boldsymbol{\mu'}}(s)= U^{\boldsymbol{\mu}}(s)$, $\forall s\in\ccalS$, holds if $\boldsymbol\mu=\boldsymbol\mu'=\boldsymbol\mu^*$, where $\boldsymbol\mu^*$ is the optimal policy \cite{rlbook}.
%
%
%
%
%

\vspace{-0.3cm}
\subsection{Myopic Effect of Biased Exploration}\label{sec:myopic}
In this section, we demonstrate the myopic benefit of the biased exploration; the proofs can be found in Appendix \ref{sec:proof1}. To formally describe it we introduce first the following definitions. Let $s_{t}=(x_{t},q_{t})$ be the PMDP state at the current \textcolor{black}{time step} $t$ of an RL episode of Algorithm \ref{alg:RL-LTL}.  Also, let $\ccalR(x_{t})\subseteq\ccalX$ denote a set collecting all MDP states that can be reached within one hop from $x_{t}$, i.e.,
\begin{equation}\label{eq:reachX}
    \ccalR(x_{t}) = \{x\in\ccalX~|~\exists a\in\ccalA(x)~ \text{such that}~\hat{P}(x_{t},a,x)>0\}.\footnote{The reachable set in \eqref{eq:reachX} is a subset of the actual set of one-hop neighbors of $x_{t}$ since \eqref{eq:reachX} uses the estimated transition probabilities \eqref{eq:probEst}.}
\end{equation}
Then, we can define the set $\ccalX_{\text{closer}}$ that collects all MDP states that are one hop reachable from $x_{t}$ and they are closer to $\ccalX_{\text{goal}}(x_{t})$ than $x_{t}$ is, i.e.,
\begin{equation}
    \ccalX_{\text{closer}}(x_{t})=\{x\in\ccalR(x_{t})~|~J_{x,\ccalX_{\text{goal}}}=J_{x_{t},\ccalX_{\text{goal}}}-1)\}.
\end{equation}
The following result shows that the probability of $x_{t+1}\in\ccalX_{\text{closer}}(x_{t})$ is higher when biased exploration is employed. 

\begin{prop}\label{prop:sampleEf1}
\textcolor{black}{Let $s_{t}=(x_{t},q_{t})$ be the PMDP state at the current \textcolor{black}{time step} $t$ of an RL episode of Algorithm \ref{alg:RL-LTL}. 
Let also $x_b\in\ccalX_{\text{closer}}(x_{t})$ denote the MDP state towards which the action $a_b$ is biased. If $\delta_b>0$ and \eqref{eq:AsAvP} holds,
\begin{equation}\label{eq:AsAvP}
    P(x_{t},a_b,x)\geq \max_{\bar{x}\in\ccalX_{\text{closer}}(x_{t})}\sum_{a}\frac{P(x_{t},a,\bar{x})}{|\ccalA(x_{t})|}, \forall x\in\ccalX_{\text{closer}}(x_{t}),
\end{equation}
where the summation is over $a\in\ccalA(x_{t})$, then we have that
\begin{equation}\label{eq:compSample}
    \mathbb{P}_b(x_{t+1}\in\ccalX_{\text{closer}}(x_t))\geq \mathbb{P}_g(x_{t+1}\in\ccalX_{\text{closer}}(x_{t})).
\end{equation}
In \eqref{eq:compSample}, $\mathbb{P}_g(x_{t+1}\in\ccalX_{\text{closer}}(x_t))$ and $\mathbb{P}_b(x_{t+1}\in\ccalX_{\text{closer}}(x_t))$ denote the probability of reaching any state $x_{t+1}\in\ccalX_{\text{closer}}(x_{t})$ starting from $x_{t}$ without and with bias introduced in the exploration phase, respectively.$\hfill\Box$}
\end{prop}

\textcolor{black}{Next, we provide a `weaker' result which, however, does not require the strong requirement of \eqref{eq:AsAvP}. The following result  shows that the probability that the next state $x_{t+1}$ will be equal to  $x_b\in\ccalX_{\text{closer}}$ (as opposed to any state in $\ccalX_{\text{closer}}$ in Prop. \ref{prop:sampleEf1}) is greater when bias is introduced in the exploration phase. 
\begin{prop}\label{prop:sampleEf2}
Let $s_{t}=(x_{t},q_{t})$ be the PMDP state at the current time step $t$ of an RL episode of Algorithm \ref{alg:RL-LTL}. 
Let also $x_b\in\ccalX_{\text{closer}}(x_{t})$ denote the MDP state towards which the action $a_b$ is biased. If $\delta_b>0$, then
\begin{equation}\label{eq:compSample2}
    \mathbb{P}_b(x_{t+1}=x_b)\geq \mathbb{P}_g(x_{t+1}=x_b),
\end{equation}
where $\mathbb{P}_g(x_{t+1}=x_b)$ and $\mathbb{P}_b(x_{t+1}=x_b)$ denote the probability of reaching at $t+1$ the state $x_b$ starting from $x_{t}$ without and with bias introduced in the exploration phase, respectively.
\end{prop}
}


\vspace{-0.3cm}
\subsection{Non-Myopic Effect of Biased Exploration}\label{sec:nonmyopicBias}

In this section, we demonstrate the non-myopic effect of the biased exploration; the proofs can be found in Appendix \ref{sec:proof2}.  
\textcolor{black}{
To present our main results, we need to introduce the following definitions. Let $s_{t}=(x_{t},q_{t})$ be the current PMDP state. Also, let $t^*=J_{x_t,\ccalX_{\text{goal}}}$ denote the length (i.e., the number of hops/MDP transitions) of the paths $p_j^t$. Recall that all paths $p_j^t$, $j\in\ccalJ$, share the same length, in terms of number of hops, by construction. 
Second, 
we define a function $\beta:\ccalJ\rightarrow [0,1]$ that maps every path $p_j^t$, $j\in\ccalJ$, into $[0,1]$ as follows:
\begin{align}\label{eq:Pj}
&\beta(p_j^t)=\prod_{m=0}^{t^*-1}\{P(x_{t+m},a_b,x_{t+m+1})\delta_b+\nonumber\\&P(x_{t+m},a^*,x_{t+m+1})(1-\epsilon)+\frac{\delta_e}{|\ccalA(x_{t+m})|}\}.
\end{align}
In \eqref{eq:Pj}, we have that $x_{t+m}=p_{j}^t(m+1)$, for all $m\in\{0,\dots,t^*-1\}$ and $a_b$ is the biased action computed at state $s_{t+m}=(x_{t+m},q_{t})$
as discussed in Section \ref{sec:biasedExpl}, i.e., using the path $p_{j^*}^{t+m}$. 
}

\begin{prop}[Most Likely Path]\label{prop:resultI}
At time step $t$ of the current RL episode, let (i) $s_t=(x_t,q_t)$ be the current PMDP state; and (ii)  $p_{j^*}^t$ be the path used to design the biased action at the time step $t$. Let $R_j$ be a (Bernouli) random variable that is true if after $t^*$ time steps (i.e., at the time step $t+t^*)$, a path $p_j^t$ has been generated, for some $j\in\ccalJ$. If there exists $\delta_b$ and $\delta_e$ satisfying the following condition
\begin{equation}\label{reqResI}
    \beta(p_{j^*}^t)\geq\max_{j\in\ccalJ}\beta(p_{j}^t),
\end{equation}
then, we have that 
$\mathbb{P}_b(R_{j^*}=1)\geq \max_{j\in\ccalJ}\mathbb{P}_{b}(R_{j}=1)$, 
where $\mathbb{P}_b(R_{j^*}=1)$ and $\mathbb{P}_b(R_{j}=1)$ stand for the  probability that $R_{j^*}=1$ and $R_{j}=1$, respectively, if the MDP evolves as per the proposed  $(\epsilon,\delta)$-greedy policy. $\hfill\Box$ 
\end{prop}


%
%
%
%
%
%

\begin{rem} [Prop. \ref{prop:resultI}]\label{rem:resI}
    \textcolor{black}{Prop. \ref{prop:resultI} implies that there exists $\delta_b$ and $\delta_e$ such that among all paths $p_j^t$, $j\in\ccalJ$, designed at time $t$, the most likely path that the MDP will generate over the next $t^*$ time steps is $p_{j^*}^t$. For instance, if $\delta_b=1$, and, therefore, $\epsilon=\delta_e=0$, then we get that \eqref{reqResI} is equivalent to $\prod_{m=0}^{t^*-1} P(x_{t+m},a_b,x_{t+m+1})\geq\max_{j\in\ccalJ}\prod_{m=0}^{t^*-1}P(\bar{x}_{t+m},\bar{a}_b,\bar{x}_{t+m+1})$, due to \eqref{eq:optUnc}-\eqref{eq:a_b}, where  $x_{t+m}=p_{j^*}^t(m+1)$,  $\bar{x}_{t+m}=p_{j}^t(m+1)$ for all $m\in\{0,\dots,t^*-1\}$, and $a_b$ and $\bar{a}_b$ denote the biased action at states $x_{t+m}$ and $\bar{x}_{t+m}$ using the path $p_{j^*}^{t+m}$. 
    }
\end{rem}

%
%

\textcolor{black}{In what follows, we show that there exists $\delta_b$ and $\delta_e$ that ensure that the probability of generating the path $p_{j^*}^t$ under the $(\epsilon,\delta)$-greedy policy (captured by $\mathbb{P}_b(R_{j^*}=1)$) is larger than the probability of generating any path $p_{j}^t$, $j\in\ccalJ$, under the $\epsilon$-greedy policy. To make this comparative analysis meaningful, hereafter, we assume that probability of exploration $\epsilon=\delta_b+\delta_e$ is the same for both policies; thus, the probability of selecting the greedy action is the same for both policies, as well. Recall again that the $\epsilon$-greedy policy can be recovered by removing bias from the $(\epsilon,\delta)$-greedy policy, i.e., by setting $\delta_b=0$.} \textcolor{black}{To present this result, we need to define a function $\eta:\ccalJ\rightarrow [0,1]$ mapping every path $p_j^t$, $j\in\ccalJ$, into $[0,1]$ as follows:
%
%
\begin{align}\label{eq:Pjg}
&\eta(p_{j}^t)=\prod_{m=0}^{t^*-1}\{P(x_{t+m},a^*,x_{t+m+1})(1-\epsilon)+\frac{\epsilon}{|\ccalA(x_{t+m})|}\}.
\end{align}
In \eqref{eq:Pjg}, we have that $x_{t+m}=p_{j}^t(m+1)$, for all $m\in\{0,\dots,t^*-1\}$ and $a^*$ is the greedy action computed at state $s_{t+m}=(x_{t+m},q_t)$. 
%
%
}

\begin{prop}[Random vs Biased Exploration]\label{thm:resultII}
At time step $t$ of the current RL episode, let (i) $s_t=(x_t,q_t)$ be the current product state; and (ii)  $p_{j^*}^t$ be the path used to design the current biased action. Let $R_{j}$ be a (Bernouli) random variable that is true if after $t^*$ time steps (i.e., at the time step $t+t^*)$, a path $p_j^t$ has been generated for some $j\in\ccalJ$ under a policy $\boldsymbol\mu$. If there exists $\delta_b$ and $\delta_e$ satisfying the following condition
\begin{equation}\label{reqResII}
    \beta(p_{j^*}^t)\geq \max_{j\in\ccalJ} \eta(p_{j}^t) 
\end{equation}
then, we have that 
$\mathbb{P}_b(R_{j^*}=1)\geq \max_{j\in\ccalJ}\mathbb{P}_{g}(R_{j}=1)$, 
where $\mathbb{P}_b(R_{j^*}=1)$ and $\mathbb{P}_g(R_{j}=1)$ stand for the probability that $R_{j^*}=1$ and $R_{j}=1$, if the MDP evolves as per the proposed  $(\epsilon,\delta)$-greedy and $\epsilon$-greedy policy, respectively. $\hfill\Box$ 
\end{prop}


\begin{rem}[Prop. \ref{thm:resultII}]\label{rem:resultII}
    \textcolor{black}{Prop. \ref{thm:resultII} states that among all paths $p_j^t$ of length $t^*$, $j\in\ccalJ$, there exists values for $\delta_b$ and $\delta_e$ under which there exists an MDP path (the one with index $j^*$) that is more likely to be generated over the next $t^*$ time steps under the $(\epsilon,\delta)$-greedy than any path $p_j^t$, $j\in\ccalJ$ that can be generated under the $\epsilon$-greedy policy. For instance, if $\delta_b=1$ and $\delta_e=0$, (i.e., $\epsilon=1$) then \eqref{reqResII} is equivalent to $\prod_{m=0}^{t^*-1} P(x_{t+m},a_b,x_{t+m+1})\geq \max_{j\in\ccalJ}\prod_{m=0}^{t^*-1}\frac{1}{|\ccalA(\bar{x}_{t+m})|}$, where $x_{t+m}=p_{j^*}^t(m+1)$, and $\bar{x}_{t+m}=p_{j}^t(m+1)$ for all $m\in\{0,\dots,t^*-1\}$. Let $A_{\text{min}}=\min_{x\in\ccalX}|\ccalA(x)|$. Then, for $\delta_b=1$, the result in Proposition \ref{thm:resultII} holds if  $\prod_{m=0}^{t^*-1} P(x_{t+m},a_b,x_{t+m+1})\geq (\frac{1}{A_{\text{min}}})^{t^*}$. The latter is true if e.g.,  $P(x_{t+m},a_b,x_{t+m+1})\geq \frac{1}{A_{\text{min}}}$ for all $m\in\{0,\dots,t^*-1\}$. We note that a similar result is presented in \cite{kantaros2020stylus} which employs a similar biased exploration to address deterministic temporal logic planning problems (see Remark 4.5 in \cite{kantaros2020stylus}).}
\end{rem}

\textcolor{black}{Proposition \ref{thm:resultII} compares the sample-efficiency of $(\epsilon,\delta)$-greedy and $\epsilon$-greedy policies with respect to a specific path $p_{j^*}^t$. In the following result, building upon Proposition \ref{thm:resultII}, we provide a more general result. Specifically, we show that the probability that after $t^*+1$ time steps a PMDP state $s=(x,q)$, where $q\in\ccalQ_{\text{goal}}$ (see \eqref{eq:Qnext}), will be reached is higher when bias is introduced in the exploration phase. We emphasize again that given the current PMDP state $s_t=(x_t,q_t)$ in an RL episode, the earliest that a PMDP state $s=(x,q)$, where $q\in\ccalQ_{\text{goal}}$ can be reached is after $t^*+1$ where $t^*=J_{x_t,\ccalX_{\text{goal}}}$ iterations. The reason is that the length of the shortest path from $x_t$ to states $\ccalX_{\text{goal}}$ that can enable the transition from $q_t$ to  $\ccalQ_{\text{goal}}$ is $t^*=J_{x_t,\ccalX_{\text{goal}}}$.}


\begin{prop}[Sample Efficiency]\label{thm:resultIII}
Let $s_t=(x_t,q_t)$ be the product state reached at the $t$-th \textcolor{black}{time step} of the current RL episode. A  state $s_{\text{goal}}=(x,q_{\text{goal}})$, where $q_{\text{goal}}\in\ccalQ_{\text{goal}}$ can be reached after at least $t^*+1$ time steps, where $t^*=J_{x_t,\ccalX_{\text{goal}}}$. 
%
%
If there exist $\delta_b$ and $\delta_e$ satisfying the following condition:
 \begin{align}\label{eq:reqResIII}
\sum_{j\in\ccalJ}\beta(p_{j}^t)\geq \sum_{j\in\ccalJ}\eta(p_{j}^t),
%
\end{align}
where $j^*$ stands for the index to the path selected as per \eqref{eq:optUnc}, then $\mathbb{P}_b(q_{t+t^*+1}\in\ccalQ_{\text{goal}})\geq \mathbb{P}_g(q_{t+t^*+1}\in\ccalQ_{\text{goal}})$, where $\mathbb{P}_b(q_{t+t^*+1}\in\ccalQ_{\text{goal}})$ and $\mathbb{P}_g(q_{t+t^*+1}\in\ccalQ_{\text{goal}})$ stand for the  probability that a PMDP state with a DRA state in $\ccalQ_{\text{goal}}$ will be reached after exactly $t^*+1$ time steps using the $(\epsilon,\delta)$-greedy and $\epsilon$-greedy policy, respectively. $\hfill\Box$
%
\end{prop}




\begin{rem}[Selecting parameters $\delta_b$ and $\delta_e$]\label{rem:deltas}
\textcolor{black}{
\textbf{(i)} The result in Proposition \ref{thm:resultIII} shows that there exist $\delta_b$ and $\delta_e$ to potentially improve sample efficiency compared to uniform/random exploration. However, selection of $\delta_b$ and $\delta_e$ as 
per Proposition \ref{thm:resultIII} requires knowledge of the actual MDP transition probabilities along all paths $p_j^t$, $j\in\ccalJ$ which are not available. To address this, the estimated transition probabilities, computed in \eqref{eq:probEst}, can be used instead. To mitigate the fact that the initial estimated probabilities may be rather inaccurate, $\delta_e$ can be selected so that $\delta_e>\delta_b$ for the first few episodes. Intuitively, this allows to initially perform random exploration to learn an accurate enough MDP transition probabilities across all directions.
%
%
Once this happens and given $\epsilon$, values for $\delta_b$ and $\delta_e$ that satisfy the requirement \eqref{eq:reqResIII} (using the estimated probabilities) can be computed by applying a simple line search algorithm over all possible values for $\delta_b\in\{0,\epsilon\}$, since $\delta_e+\delta_b=\epsilon$. \textbf{(ii)} A more efficient approach would be to pick $\delta_b$ based on Proposition \ref{thm:resultII} instead of \ref{thm:resultIII}. The reason is that searching for $\delta_b$ that satisfies \eqref{reqResII} requires less computations than \eqref{eq:reqResIII}; see also Remark \ref{rem:resultII}. 
\textbf{(iii)} An even more computationally efficient, but heuristic, approach to pick $\delta_b$ and $\delta_e$ is the following. We select $\delta_b$ and $\delta_e$ so that $\delta_e>\delta_b$ for the first few episodes to learn an accurate enough MDP model and then allow $\delta_e<\delta_b$ to prioritize exploration towards directions that may contribute to mission progress while letting both $\delta_b$ and $\delta_b$ to asymptotically converge to $0$. 
%
Nevertheless, the values for $\delta_b$ and $\delta_e$ selected in this way may not satisfy the requirements mentioned in Propositions \ref{prop:resultI}, \ref{thm:resultII}, and \ref{thm:resultIII}.}
\end{rem}

\begin{rem} [Limitations] \textcolor{black} {Alternative definitions of $d_F$ may affect the construction of the set $\ccalQ_{\text{goal}}$ in \eqref{eq:Qnext}. Currently, $d_F$ captures the shortest path, in terms of number of hops, between a DRA state and the set of accepting states. However, this definition neglects the underlying MDP structure which may compromise sample-efficiency. Specifically, the shortest DRA-based path may be harder for the MDP to realize than a longer DRA-based path, depending on the MDP transition probabilities. The result presented in Proposition \ref{thm:resultIII} shows that given a distance function $d_F$ and, consequently, $\ccalQ_{\text{goal}}$,  there exist conditions that the parameters $\delta_b$ and $\delta_e$ should satisfy, so that the probability of reaching $\ccalQ_{\text{goal}}$ within the minimum possible number of time steps (i.e., $J_{x,\ccalX_{\text{goal}}}$ time steps) is larger when the $(\epsilon,\delta)$-greedy policy is used. This does not necessarily imply that the probability of eventually reaching accepting states is also larger as this depends on the definition of $d_F$ and, consequently, $\ccalQ_{\text{goal}}$. Designing $d_F$ that optimizes sample-efficiency is a future research direction. 
%
However, our comparative experiments in Section \ref{sec:Sim} demonstrate sample-efficiency of the proposed method under various settings.}
%
%
\end{rem}

%% file: files/sim.tex
To demonstrate the sample-efficiency of our method, we provide extensive comparisons against existing model-free and model-based RL algorithms. All methods have been implemented on Python 3.8 and evaluated on a computer with an Nvidia RTX 3080 GPU, 12th Gen Intel(R) Core(TM) i7-12700K CPU, and 8GB RAM.

\vspace{-0.3cm}
\subsection{Setting up Experiments \& Baselines}
\textbf{MDP:} We consider environments represented as $10\times 10$, $20\times 20$, and $50\times50$ discrete grid worlds,
%
resulting in MDPs with $|\ccalX|=100, 400$, and $2,500$ states denoted by $\mathfrak{M}_1$, $\mathfrak{M}_2$, and $\mathfrak{M}_3$, respectively. 
The robot has nine actions:`left', `right', `up', `down', `idle' as well as the corresponding four diagonal actions. At any MDP state $x$, excluding the boundary ones, the set of actions $\ccalA(x)$ that the robot can apply includes eight of these nine actions that are randomly selected while ensuring that
the idle action is available at any state. The set of actions at boundary MDP states exclude those ones that drive the robot outside the environment.
The transition probabilities are designed so that given any action, besides `idle', the probability of reaching the intended state is randomly selected from the interval $[0.7,0.8]$ while the probability of reaching neighboring MDP states is randomly selected as long as the summation of transition probabilities over the next states $x'$ is equal to $1$, for a fixed action $a$ and starting state $x$.
The action `idle' is applied deterministically. 

\begin{figure}[t]
    \centering
     \includegraphics[width=1\linewidth]{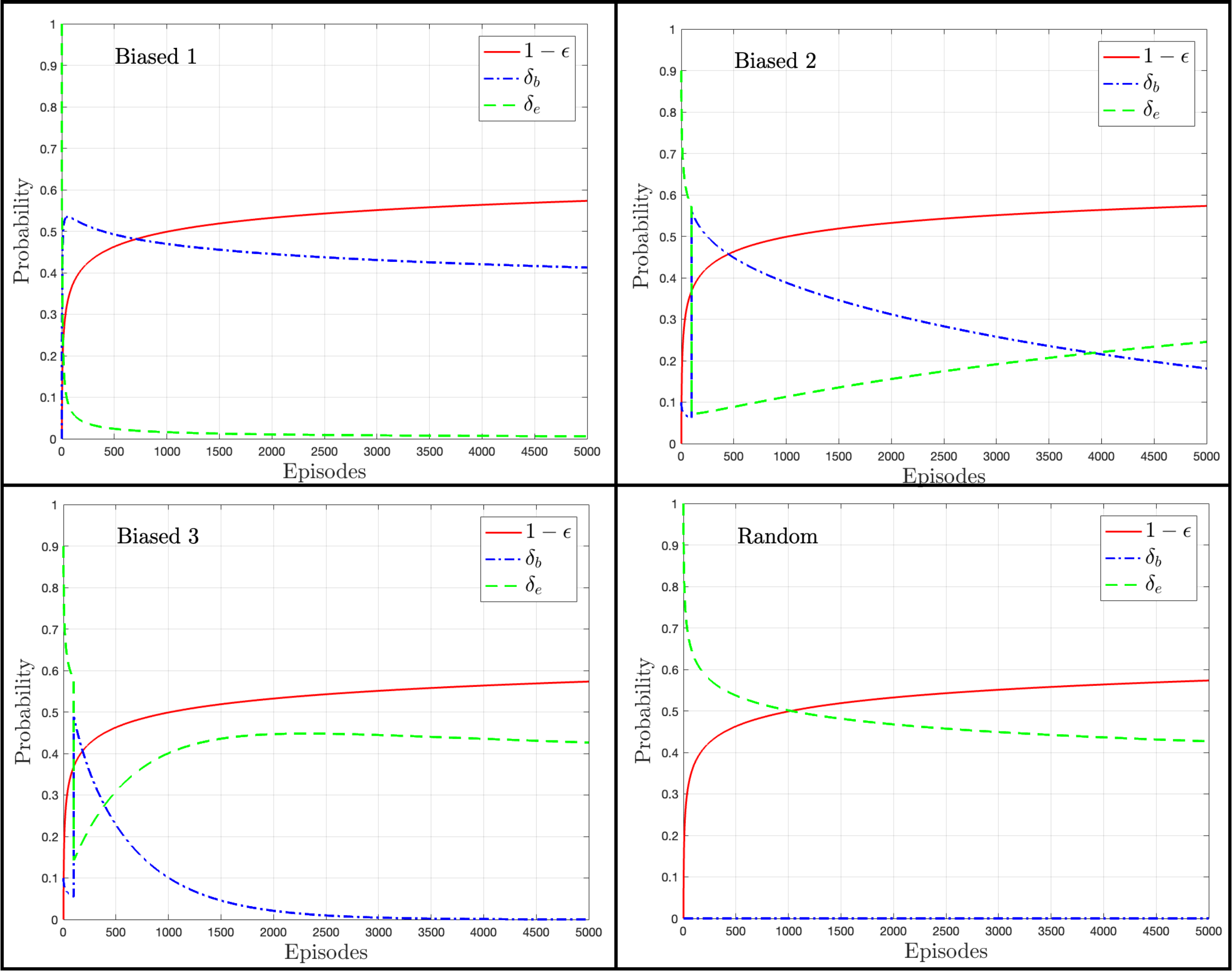}
     \vspace{-0.6cm}
    \caption{Decay rates of the parameters $\delta_e$, $\delta_b$, and $\epsilon$ considered in Section \ref{sec:Sim} for $\mathfrak{M}_1$ and $\mathfrak{M}_2$. 
    The rate at which $1-\epsilon$ (red) increases is the same in all figures. 
    As the number of episodes goes to infinity, $1-\epsilon$ converges to $1$ and both $\delta_b$ and $\delta_e$ converge to $0$.
    Notice that, in the bottom right figure, $\delta_b$ is always equal to $0$ resulting in random exploration ($\epsilon$-greedy policy).  }\vspace{-0.7cm}
    \label{fig:decayRates}
\end{figure}

\textbf{Baselines:} In the following case studies we demonstrate the performance of Algorithm \ref{alg:RL-LTL} when it is equipped with the proposed $(\epsilon,\delta)$-greedy policy \eqref{eq:policy}, the $\epsilon$-greedy policy, the Boltzman policy, and the UCB1 policy. Notice that Alg. \ref{alg:RL-LTL} is model-free when it is equipped with these baselines as it does not require learning the MDP. We also compare it against a standard model-based approach that explicitly computes and stores the product MDP (PMDP) \cite{baier2008principles}. Computing the PMDP requires learning the underlying MDP model which can be achieved e.g., by simply letting the agent randomly explore the environment and then estimating the transition probabilities as in \eqref{eq:probEst}.\footnote{This would result in learning transition probabilities of  $\mathfrak{M}_1$ 
and $\mathfrak{M}_2$ in $1.1$ and $90$ minutes, respectively, with maximum error equal to $0.05$.} In our implementation, we directly use the ground-truth MDP transition probabilities giving an `unfair' advantage to the model-based approach over the proposed one. Given the resulting PMDP, we apply dynamic programming to compute the optimal policy and its satisfaction probability \cite{baier2008principles}.


\begin{figure*}[t]
    \centering
    \includegraphics[width=1\linewidth]{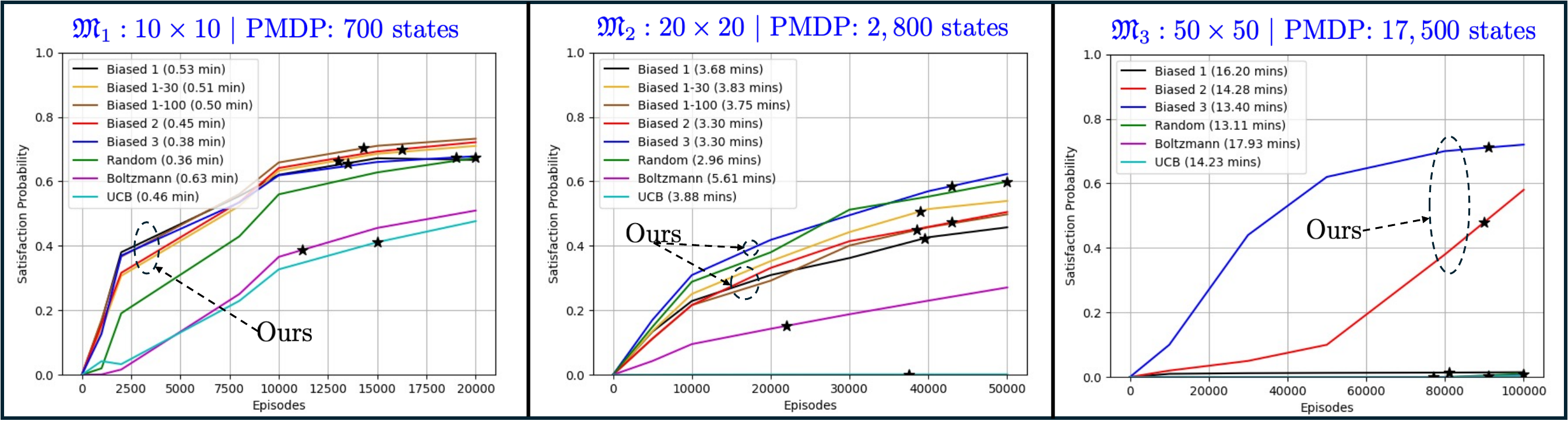}\vspace{-0.3cm}
    \caption{A Simple Coverage Task (Section \ref{sec:coverage}): Comparison of average satisfaction probability $\bar{\mathbb{P}}$ when Algorithm \ref{alg:RL-LTL} is applied with the proposed $(\epsilon,\delta)$-greedy policy, $\epsilon$-greedy policy, Boltzmann policy, and UCB1 policy over the MDPs $\mathfrak{M}_1$, $\mathfrak{M}_2$, and $\mathfrak{M}_3$. 
    $\text{Biased 1}-30$ and $\text{Biased 1}-100$ refer to the cases where the Biased 1 exploration method is applied under the constraint that the MDP transition probabilities are updated only during the first $30$ and $100$ episodes, respectively. The legend also includes the total runtime per method. The black stars on top of each reward curve denote the training episode where the corresponding policy is when the fastest policy has finished training over the total number of episodes. 
    }\vspace{-0.7cm}
    \label{fig:coverage}
\end{figure*}

To examine sensitivity of the proposed algorithm with respect to the parameters  $\delta_e$ and $\delta_b$, 
we have considered three different decay rates for $\delta_e$ and $\delta_b$, as per (iii) in Remark \ref{rem:deltas}. Hereafter, we refer to the corresponding exploration strategies as `Biased 1', `Biased 2', and `Biased 3', and `Random', where the latter corresponds to the $\epsilon$-greedy policy. The rate at which $\delta_b$ decreases over time gets smaller as we proceed from `Biased 1', `Biased 2', `Biased 3', to `Random'. In other words, `Biased 1' incurs the most 'aggressive' bias in the exploration phase.
The evolution of these parameters for the MDPs $\mathfrak{M}_1$ and $\mathfrak{M}_2$ is illustrated in Fig. \ref{fig:decayRates}. Similar biased strategies were selected for $\mathfrak{M}_3$. The only difference is that $\delta_b$ is designed so that it converges to $0$ slower due to the larger size of the state space. The corresponding mathematical formulas are provided in Appendix \ref{sec:decayRates}.
To make the comparison between the $(\epsilon,\delta)$- and the $\epsilon$-greedy policy fair, we select the same $\epsilon$ for both. 
%
%
The Boltzmann control policy is defined as follows: $\boldsymbol\mu_B(s) = \frac{e^{Q^{\boldsymbol\mu_B}(s,a)/T}}{\sum_{a'\in\ccalA_{\mathfrak{P}}}e^{Q^{\boldsymbol\mu_B}(s,a')/T}}$, 
%
where $T\geq 0$ is the temperature parameter. 
%
The UCB1 control policy is defined as:
$\boldsymbol\mu_U(s) = \argmax_{a\in\ccalA_{\mathfrak{P}}}\left[Q^{\boldsymbol\mu_U}(s,a) + C\times \sqrt{\frac{2\log(N(s))}{n(s,a)}}\right]$, where (i) $N(s)$ and $n(s,a)$ denote the number of times state $s$ has been visited and the number of times action $a$ has been selected at state $s$ and (ii) $C$ is an exploration parameter. 
This control policy is biased towards the least explored directions. In each case study, we pick values for $C$ and $T$ from a fixed set that yield the best performance. 
In all case studies, we adopt the reward function in \eqref{eq:RewardQ} with $\gamma=0.99$ and $r_{\ccalG}=10$, $r_{\ccalB}=-0.1$, $r_{d}=-100$, and $r_o=0$. To convert the LTL formulas into DRA, we have used the ltl2dstar toolbox \cite{ltl2dstar}.
%

\begin{figure*}
\centering
\includegraphics[width=1\linewidth]{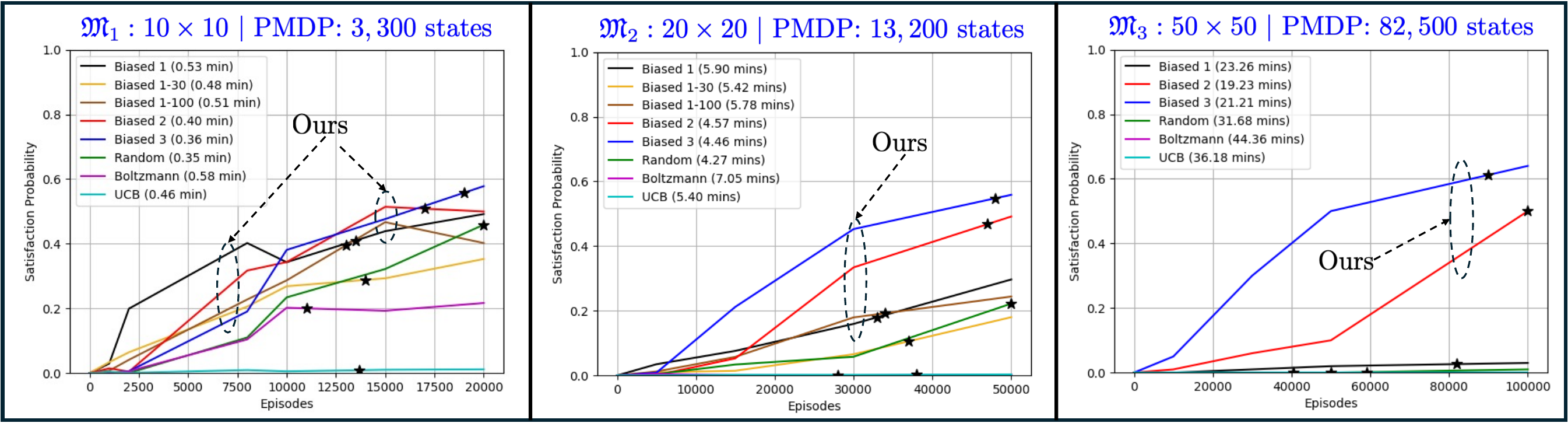}\vspace{-0.3cm}
 \caption{A More Complex Coverage Task (Section \ref{sec:reachI}): Comparison of average accumulated reward (top row) and satisfaction probability $\bar{\mathbb{P}}$ (bottom row) when Algorithm \ref{alg:RL-LTL} is applied with the proposed $(\epsilon,\delta)$-greedy policy, $\epsilon$-greedy policy, Boltzmann policy, and UCB1 policy over the MDPs $\mathfrak{M}_1$, $\mathfrak{M}_2$, and $\mathfrak{M}_3$. 
 The legend also includes the total runtime per method. The black stars on top of each reward curve denote the training episode where the corresponding policy is when the fastest policy has finished training over the total number of episodes.
 } \vspace{-0.6cm}
 \label{fig:task1}
\end{figure*}

\textbf{Performance Metrics:} 
 %
%
We utilize the satisfaction probabilities of the policies learned at various stages during training to assess performance of our algorithm and the baselines. Specifically, given a learned/fixed policy $\boldsymbol{\mu}$ and an initial PMDP state $s=(x,q_D^0)$, we compute the probability $\mathbb{P}(\boldsymbol{\mu}\models \phi|s=(x,q_D^0))$ 
using dynamic programming. 
%
%
We compute this probability for all $x\in\ccalX$ and then we compute the average satisfaction probability $\bar{\mathbb{P}}=[\sum_{\forall x\in\ccalX}\mathbb{P}(\boldsymbol{\mu}\models \phi|s=(x,q_D^0))]/{|\ccalX|}$. We report the average $\bar{\mathbb{P}}$ over five runs; see Figs. \ref{fig:coverage}-\ref{fig:taskDisjoint}. The satisfaction probabilities are computed using the unknown-to-the-agent MDP transition probabilities.
Since runtimes for a training episode may differ across methods, we also report runtime metrics; see Figs. \ref{fig:coverage}-\ref{fig:taskDisjoint}. Specifically, we document the runtimes required for all methods to complete a a predetermined maximum number of episodes, as well as the training episode each method reaches when the fastest one completes the training process. This allows us to compare satisfaction probabilities over the policies more fairly based on fixed runtimes rather than a fixed number of training episodes. 
%
%

%

\textbf{Summary of Comparisons:}{ 
Our experiments show that the proposed $(\epsilon,\delta)$-greedy policy outperforms the model-free baselines, learning policies with higher satisfaction probabilities over the same timeframe. This performance gap widens significantly as the size of the PMDP increases. Specifically, our method begins learning policies with non-zero satisfaction probabilities within the first few hundred training episodes. The baselines can catch up relatively quickly, narrowing the performance gap, typically after a few thousand episodes, but only in small PMDPs (fewer than $10,000$ states). In larger PMDPs (more than $10,000$ states), our method significantly outperforms the model-free baselines.
Additionally, the proposed $(\epsilon,\delta)$-greedy policy and the $\epsilon$-greedy policy have similar runtimes, while they tend to be faster than UCB and, especially, Boltzmann. 
The model-based approach, on the other hand, demonstrates faster computation of the optimal policy compared to model-free baselines, including ours, when applied to small PMDPs (e.g., with fewer than $5,000$ states). However, this approach is memory inefficient, requiring storage of the PMDP and the action value function $Q^{\boldsymbol{\mu}}$. As a result, it failed to handle case studies with large PMDPs (more than $15,000$ states). In contrast, our method was able to handle PMDPs with hundreds of thousands of states; see e.g., Section \ref{sec:twoRob}. 
}

\begin{rem}[Limitations \& Implementation Improvements]
A limitation of our method compared to model-free baselines is that it requires learning an MDP model, which can become memory-inefficient over large-scale MDPs. 
However, we believe that this limitation can
be mitigated by more efficient implementations of our approach. For instance, in our current implementation \cite{codeAccRL}, we store all learned MDP transition probabilities used to compute the biased action. However, the selection of the biased action does not require learning all transition probabilities; see \eqref{eq:a_b}. Instead, it only requires learning which action is most likely to drive the system from a state $x$ to a neighboring state $x'$. Once this property is learned for a pair of states $x$ and $x'$, the estimated transition probabilities $\hat{P}(x,a,x')$ in \eqref{eq:probEst} can be discarded.
\end{rem}

\vspace{-0.3cm}
\subsection{Case Study I: A Simple Coverage Task}\label{sec:coverage}
\vspace{-0.1cm}
First, we consider a coverage/sequencing mission requiring the agent to eventually reach the states $99$, and $46$ or $90$ while avoiding $99$ until $33$ is reached, and always avoiding the obstacle states $73$, $24$, $15$, and $88$. This task is captured by the following LTL formula $\phi=(\Diamond \pi^{99})\wedge \Diamond (\pi^{46}\vee \pi^{90})\wedge (\Diamond \pi^{33})\wedge (\neg \pi^{99}\ccalU \pi^{33})\wedge \square \neg \pi^{\text{obs}}$, where $\pi^{\text{obs}}$ is satisfied when the robot visits one of the obstacle states. This formula corresponds to a DRA with $7$ states and $1$ accepting pair. Thus, the PMDP constructed using $\mathfrak{M}_1$, $\mathfrak{M}_2$, and $\mathfrak{M}_3$ has $700$, $2,800$, and $17,500$ states, respectively.

The comparative results are shown in Fig. \ref{fig:coverage}. Our algorithm achieves the best performance when applied to $\mathfrak{M}_1$, regardless of the biased strategy. As for $\mathfrak{M}_2$, the best performance is achieved by our $(\epsilon,\delta)$-greedy policy coupled with 'Biased 3', followed closely by the $\epsilon$-greedy policy, 'Biased 2', and 'Biased 1'. 
Notice that $\epsilon$-greedy can catch up quickly when applied to $\mathfrak{M}_1$ and $\mathfrak{M}_2$ due to the relatively small size of the resulting PMDPs. 
This figure also shows the performance of 'Biased 1' when the MDP transition probabilities are updated only for the first $30$ and $100$ episodes for $\mathfrak{M}_1$ and $\mathfrak{M}_2$. The performance of our approach is not significantly affected by this choice, demonstrating robustness against model inaccuracies. This occurs because our algorithm does not require learning the ground truth values of the transition probabilities to compute the biased action; see \eqref{eq:a_b}. 

The benefit of our method becomes more pronounced when $\mathfrak{M}_3$ is considered, resulting in a larger PMDP. In this case, the average satisfaction probability $\bar{\mathbb{P}}$ of the policies learned by all baselines is close to $0$ after $100,000$ training episodes (approximately 15 minutes). In contrast, the proposed $(\epsilon,\delta)$-greedy policy, coupled with `Biased 2' and `Biased 3', learned policies with $\bar{\mathbb{P}}=0.58$ and $\bar{\mathbb{P}}=0.71$, respectively, within the same timeframe. Also, notice that `Biased 1' failed to yield a satisfactory policy for $\mathfrak{M}_3$ within the same timeframe; recall that `Biased 1' for $\mathfrak{M}_3$ is more aggressive than `Biased 1 for $\mathfrak{M}_1$ and $\mathfrak{M}_2$. We attribute this to the aggressive nature of this exploration strategy towards a desired high-reward path, which possibly does not allow the agent to sufficiently explore a significant portion of the PMDP state space, resulting in a low average satisfaction probability. This shows that increasing the amount of bias does not necessarily yield policies with higher satisfaction probabilities. 


The model-based approach was able to compute the optimal policy for the MDPs $\mathfrak{M}_1$ and $\mathfrak{M}_2$, but failed for $\mathfrak{M}_3$ due to excessive memory requirements. Specifically, the optimal policy for $\mathfrak{M}_1$ was computed in $0.5$ minutes, and for $\mathfrak{M}_2$, it took $5.98$ minutes (without including the time to learn the MDP model). The corresponding optimal average satisfaction probabilities $\bar{\mathbb{P}}$ were $0.916$ for $\mathfrak{M}_1$ and $0.911$ for $\mathfrak{M}_2$. We noticed that the model-based approach tends to be faster than the model-free baselines, particularly for smaller PMDPs. 

\vspace{-0.4cm}
\subsection{Case Study II: A More Complex Coverage Task}\label{sec:reachI}
\vspace{-0.1cm}
Second, we consider a more complex sequencing task compared to the one in Section \ref{sec:coverage}, which involves visiting a larger number of MDP states. The goal is to eventually reach the MDP states $x=81, 95, 80, 88$, and $92$ in any order, while always avoiding the states $x=5, 15, 54, 32, 24, 66, 42, 70$, and $71$ representing obstacles in the environment. This task can be formulated using the following LTL formula: $\phi=\Diamond \pi^{81} \wedge \Diamond \pi^{95} \wedge \Diamond \pi^{80} \wedge \Diamond \pi^{88} \wedge \Diamond \pi^{80} \wedge \Diamond \pi^{92} \wedge \square \neg \pi^{\text{obs}}$, where $\pi^{\text{obs}}$ is true if the robot visits any of the obstacle states. This formula corresponds to a DRA with $33$ states and $1$ accepting pair. Therefore, the PMDPs constructed using $\mathfrak{M}_1$, $\mathfrak{M}_2$, and $\mathfrak{M}_3$ have $3,300$, $13,200$, and $82,500$ states, respectively, which are significantly larger than those of Section \ref{sec:coverage}.

Overall, our method, especially when coupled with `Biased 2' and `Biased 3', learns policies with higher satisfaction probabilities faster than the baselines; see Fig. \ref{fig:task1}. The benefit of our method is more pronounced as the PMDP size increases, as shown in the cases of $\mathfrak{M}_2$ and $\mathfrak{M}_3$. For example, when considering the MDP $\mathfrak{M}_3$, our method equipped with 'Biased 2' and 'Biased 3' learns policies with $\bar{\mathbb{P}}=0.55$ and $\bar{\mathbb{P}}=0.64$, respectively, while $\bar{\mathbb{P}}<0.05$ for all other baselines, given the same amount of training time. Also, as in Section \ref{sec:coverage}, observe that `Biased 1' failed to learn a satisfactory policy for $\mathfrak{M}_3$.

The model-based baseline computed the optimal policy $\boldsymbol{\mu}^*$ for the MDPs $\mathfrak{M}_1$ and $\mathfrak{M}_2$ in $1.1$ and $112.15$ minutes, respectively, while it failed to compute the optimal policy for  $\mathfrak{M}_3$ due to excessive memory requirements.
%
The average optimal satisfaction probability of the learned policies for $\mathfrak{M}_1$ and $\mathfrak{M}_2$ is $0.9854$ and $0.9466$, respectively. 

\vspace{-0.3cm}
\begin{figure*}[t]
    \centering
     \includegraphics[width=1\linewidth]{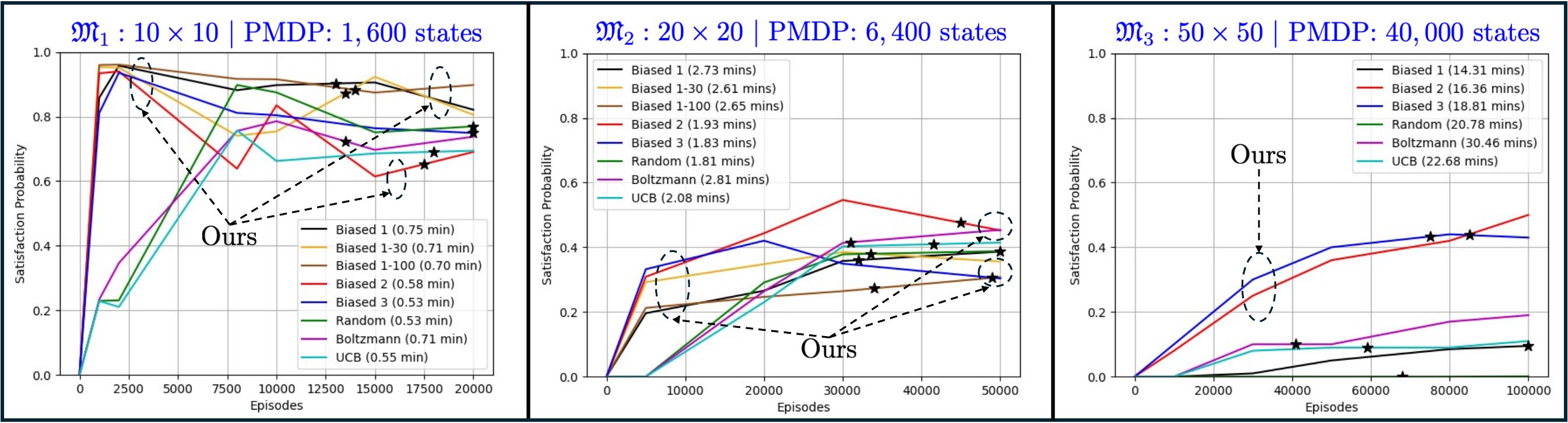}\vspace{-0.3cm}
    \caption{Surveillance Task (Section \ref{sec:singleSurv}): Comparison of average  satisfaction probability $\bar{\mathbb{P}}$ when Algorithm \ref{alg:RL-LTL} is applied with the proposed $(\epsilon,\delta)$-greedy policy, $\epsilon$-greedy policy, Boltzmann policy, and UCB1 policy over the MDPs $\mathfrak{M}_1$, $\mathfrak{M}_2$, and $\mathfrak{M}_3$. 
    The legend also includes the total runtime per method. The black stars on top of each reward curve denote the training episode where the corresponding policy is when the fastest policy has finished training over the total number of episodes. 
    }\vspace{-4mm}
    \label{fig:task3}
\end{figure*}

\begin{figure*}[t]
    \centering
     \includegraphics[width=1\linewidth]{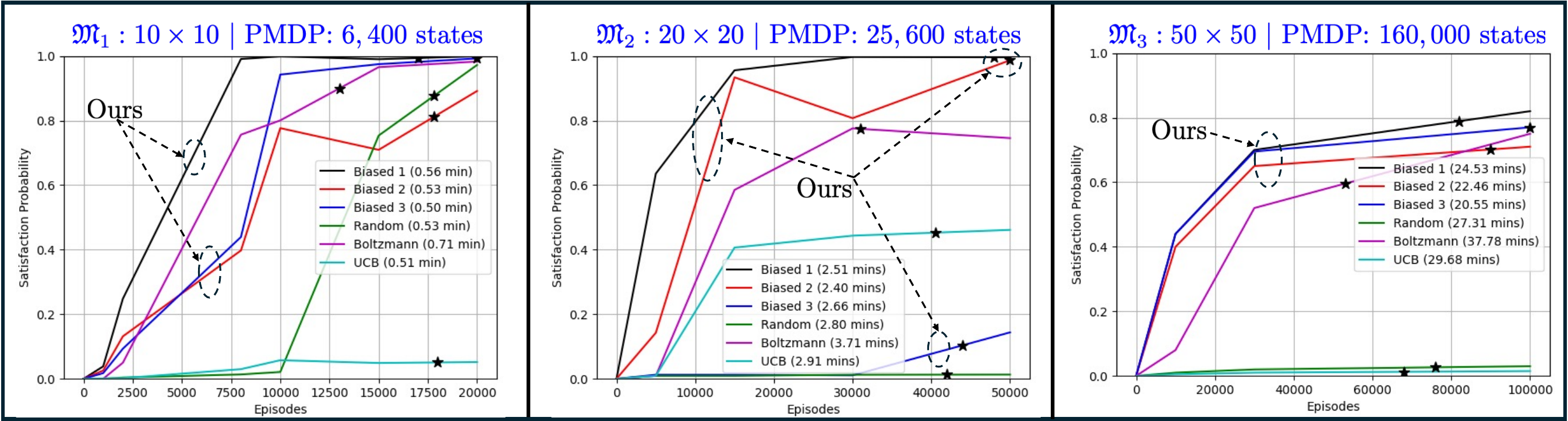}\vspace{-0.3cm}     
    \caption{Disjoint Surveillance Task (Section \ref{sec:twoRob}): Comparison of average  satisfaction probability $\bar{\mathbb{P}}$ when Algorithm \ref{alg:RL-LTL} is applied with the proposed $(\epsilon,\delta)$-greedy policy, $\epsilon$-greedy policy, Boltzmann policy, and UCB1 policy over the MDPs $\mathfrak{M}_1$, $\mathfrak{M}_2$, and $\mathfrak{M}_3$. 
    The legend includes the total runtime per method. The black stars on top of each reward curve denote the training episode where the corresponding policy is when the fastest policy has finished training over the total number of episodes. 
    }\vspace{-0.7cm}
    \label{fig:taskDisjoint}
\end{figure*}

\subsection{Case Study III: Surveillance Task}\label{sec:singleSurv}
\vspace{-0.1cm}
Third, we consider a surveillance/recurrence mission captured by the following LTL formula: $\phi=\square\Diamond \pi^{90} \wedge \square\Diamond \pi^{70} \wedge \square\Diamond (\pi^{80}\vee\pi^{63}) \wedge \square\Diamond \pi^{88} \wedge (\neg \pi^{88}\ccalU\pi^{90})\wedge \square \neg \pi^{\text{obs}}.$
This formula requires the robot to (i) visit infinitely often and in any order the states $90$, $70$, $80$ or $63$ and $88$; (ii) avoid reaching $88$ until $80$ is visited; and (iii) always avoid the obstacle in state $33$. The corresponding DRA has $16$ states and $1$ accepting pair. Thus, the PMDP constructed using $\mathfrak{M}_1$, $\mathfrak{M}_2$, and $\mathfrak{M}_3$ has $1,600$, $6,400$, and $40,000$ states, respectively.

The comparative performance results are shown in Figure \ref{fig:task3}. Observe that the $(\epsilon,\delta)$-greedy policy, especially when paired with 'Biased 2' and 'Biased 3', performs better that the model-free baselines in terms of sample-efficiency across all considered MDPs. For instance, in the case of $\mathfrak{M}_1$, our proposed algorithm learns policies with average satisfaction probabilities ranging from $0.75$ to $0.9$, depending on the biased exploration strategy, within $2,500$ training episodes. In contrast, the average satisfaction probability for the baselines is around $0.4$ after the same number of episodes. As the number of episodes increases, the baselines manage to catch up due to the relatively small PMDP size. Similar trends are observed for $\mathfrak{M}_2$. As in the other case studies, the benefit of our method becomes more evident when considering $\mathfrak{M}_3$, which yields a significantly larger PMDP. In this scenario, our proposed algorithm, coupled with 'Biased 2' and 'Biased 3', learns a control policy with satisfaction probabilities of $\bar{\mathbb{P}}=0.51$ and $\bar{\mathbb{P}}=0.45$ within $100,000$ episodes (or approximately $20$ minutes), respectively. In contrast, the baselines achieve satisfaction probabilities $\bar{\mathbb{P}}<0.2$ within the same timeframe. As discussed in Section \ref{sec:reachI}, 'Biased 1' performs poorly in $\mathfrak{M}_3$, possibly due to its aggressive bias. 


The model-based approach can compute the optimal policy only for the MDPs $\mathfrak{M}_1$ and $\mathfrak{M}_2$ while it failed in the case of $\mathfrak{M}_3$ due to excessive memory requirements. Regarding the MDP $\mathfrak{M}_1$ it computed an optimal policy corresponding to $\bar{\mathbb{P}}=0.989$ within $2.31$ minutes. As for the MDP $\mathfrak{M}_2$, it computed the optimal policy with $\bar{\mathbb{P}}=0.981$ within $ 7.71$ minutes. 

\vspace{-0.4cm}
\subsection{Case Study IV: Disjoint Task}\label{sec:twoRob}
\vspace{-0.1cm}
Finally, we consider a mission $\phi$ with two disjoint sub-tasks, i.e., $\phi=\phi_1\vee\phi_2$ requiring the robot to accomplish either $\phi_1$ or $\phi_2$. The sub-tasks are defined as 
%
%
$\phi_1=(\Diamond \pi^{99} \wedge \Diamond \pi^{45} \wedge \Diamond \pi^{32}\wedge \square \neg  \pi^{64})$ and  $\phi_2=(\Diamond \pi^{18}\wedge \Diamond \pi^{72} \wedge \Diamond \pi^{4})$. The LTL formula $\phi$ corresponds to a DRA with $64$ states and $2$ accepting pairs. As a result, the PMDP constructed using $\mathfrak{M}_1$, $\mathfrak{M}_2$, and $\mathfrak{M}_3$ has $6,400$, $25,600$, and $160,000$ states, respectively.
This task requires the robot to eventually either  visit the states $99$, $45$, and $32$ while always avoiding $64$ or visit the states $18$, $72$, and $4$. Notice that the optimal satisfaction probability of $\phi_1$ and $\phi_2$ is $1$ and less than $1$, respectively.

The comparative results are reported in Figure \ref{fig:taskDisjoint}. 
In $\mathfrak{M}_1$, our method coupled with 'Biased 1' achieves the best performance, closely followed by 'Biased 3'. Both biased exploration strategies result in a control policy satisfying $\phi$ with probability very close to $1$ in approximately $0.5$ minutes. Additionally, all other baselines, except UCB, perform satisfactorily, learning policies with $\mathbb{P}\in [0.8,0.9]$ in the same time frame. The performance gap between our method and the baselines becomes more pronounced with the larger PMDPs constructed using $\mathfrak{M}_2$ and $\mathfrak{M}_3$. 
%
%
%
Specifically, in $\mathfrak{M}_2$, `Biased 1' and `Biased 2' achieve the best performance followed by `Boltzmann', `UCB', `Biased 2', and `$\epsilon$-greedy'. In fact, `Biased 1' and `Biased 2' still manage to learn a policy with $\bar{\mathbb{P}}$ very close to $1$ in $2.40$ mins while for the other baselines it holds that $\bar{\mathbb{P}}<0.8$.
It is worth noting that the performance of `Biased 3' has dropped significantly compared to $\mathfrak{M}_1$. This drop may be attributed to $\delta_b$ converging quite fast to $0$ relative to the large size of the PMDP. 
In fact, once $\delta_b$ is almost equal to $0$, then the $(\epsilon,\delta)$-greedy policy closely resembles the standard $\epsilon$-greedy policy which in this case has also learned a policy with very low average satisfaction probability. Recall that $\mathfrak{M}_2$ shared exactly the same biased exploration strategies (`Biased 1', `Biased 2', and `Biased 3') across all case studies regardless of the PMDP size. 
However, the PMDP for $\mathfrak{M}_2$ is significantly larger than the ones considered in the other case studies which may explain the poor performance of `Biased 3' compared the other $\mathfrak{M}_2$ case studies.
Observe in $\mathfrak{M}_3$ that our method outperforms all baselines. 
Specifically, within $20.55$ mins, the average satisfaction probability corresponding to `Biased 1', `Biased 2', `Biased 3', and `Boltzmann' is  $0.8$, $0.71$, $0.78$, and $0.6$ respectively. The Boltzmann policy requires in total $37.78$ minutes to eventually yield a policy with $\bar{\mathbb{P}}=0.76$. Finally, the model-based approach was able to compute an optimal policy only for $\mathfrak{M}_1$ within $6.1$ minutes with $\bar{\mathbb{P}}=0.9772$; interestingly, model-free methods are faster in this case study.

%% file: files/extensionsLDBA.tex
In this appendix, we show that the proposed exploration strategy can be extended to Limit Deterministic B\"uchi Automaton (LDBA) that typically have a smaller state space than DRA which can further accelerate learning \cite{sickert}. 
First, any LTL formula can be converted in an LDBA  defined as follows:

%
\begin{definition}[LDBA \cite{sickert}]\label{ldbadef}
An LDBA is defined as $\mathfrak{A} = (\ccalQ, q_0, \Sigma, \ccalF, \delta)$ where $\ccalQ$ is a finite set of states, $q_0\in\ccalQ$ is the initial state, $\Sigma=2^{\mathcal{AP}}$ is a finite alphabet, $\ccalF = \{\ccalF_1,\dots, \ccalF_f \}$ is the set of accepting conditions where $\ccalF_j \subset\ccalQ$, $1\leq j \leq f$, and $\delta:\ccalQ\times\Sigma\rightarrow 2^{\ccalQ}$ is a transition relation. The set of states $\ccalQ$ can be partitioned into two disjoint sets $\ccalQ = \ccalQ_N \cup \ccalQ_D$, so that (i) $\delta(q,\pi)\subset\ccalQ_D$ and $|\delta(q,\pi)|=1$, for every state $q\in\ccalQ_D$ and $\pi\in\Sigma$; and (ii) for every $\ccalF_j\in\ccalF$, it holds that $\ccalF_j\subset\ccalQ_D$
and there are $\varepsilon$-transitions from $\ccalQ_N$ to $\ccalQ_D$. $\hfill\Box$ 
\end{definition}
An infinite run $\rho$ of $\mathfrak{A}$ over an infinite word $w=\sigma_0\sigma_1\sigma_2\dots\in\Sigma^{\omega}$, $\sigma_t\in\Sigma=2^{\mathcal{AP}}$ $\forall t\in\mathbb{N}$, is an infinite sequence of states $q_t\in\ccalQ$, i.e.,  $\rho=q_0q_1\dots q_t\dots$, such that
$q_{t+1}\in\delta(q_t,\sigma_t)$. The infinite run $\rho$ is called \textit{accepting} (and the respective word $w$ is accepted by the LDBA) if
$\texttt{Inf}(\rho)\cap\ccalF_j\neq\emptyset,\forall j\in\{1,\dots,f\},$ %
where $\texttt{Inf}(\rho)$ is the set of states that are visited infinitely often by $\rho$. 
Also, an $\varepsilon$-transition allows the automaton to change its state without reading any specific input. In practice, the $\varepsilon$-transitions between $ \mathcal{Q}_N $ and $ \mathcal{Q}_D $ reflect the ``guess'' on reaching $ \mathcal{Q}_D $: accordingly, if after an $\varepsilon$-transition the associated labels in the accepting LDBA set cannot be read, or if the accepting states cannot be visited, then the guess is deemed to be wrong, and the trace is disregarded and is not accepted by the automaton. However, if the trace is accepting, then the trace will stay in $\mathcal{Q}_D$ ever after, i.e. $\mathcal{Q}_D$ is invariant. 

Given a (non-pruned) LDBA, we construct the product MDP (PMDP), similarly to Definition \ref{def:prodMDP}. The formal definition of this PMDP can be found in \cite{bozkurt2020control,hasanbeig2019reinforcement}. 
To synthesize a policy that satisfies the LDBA accepting condition,  we can adopt any reward function for the product MDP proposed in the literature  \cite{hasanbeig2019reinforcement,bozkurt2020control}. 
%
Once the LDBA is constructed, it is pruned exactly as discussed in Section \ref{sec:prune}. The $\epsilon$-transitions are not pruned. Given the resulting automaton, similar to \eqref{eq:dist2G}, we define the distance to an accepting set of states $\ccalF_j$ as $d_F(q,\ccalF_j)=\min_{q_G\in\ccalF_j} d(q,q_G)$
where $d(q,q_G)$ is defined as in \eqref{eq:dist1}. This  function is used to bias exploration so that each set $\ccalF_j$ is visited infinitely often. 
To design a biased exploration strategy that can account for the LDBA accepting condition, we first define the set $\ccalV$ that collects indices $j$ to the set of accepting states $\ccalF_j$ that have not been visited during the current RL episode. Then, among all non-visited set of accepting states $\ccalF_j$, we pick one randomly based on which we define the set $\ccalQ_{\text{goal}}(q_t)$. Similar to \eqref{eq:Qnext}, we define the set $\ccalQ_{\text{goal}}(q_t)$ as:
$\ccalQ_{\text{goal}}(q_t)=\{q'\in\ccalQ~|~(\exists\sigma\in\Sigma_{\text{feas}}~\text{such that}~q'\in\delta(q_t,\sigma)) \wedge (d_F(q',\ccalF_j)=d_F(q_t,\ccalF_j)-1)\}$,
%
where $j\in\ccalV$. Recall, that all $\epsilon$- transitions in the LDBA are feasible. Thus, by definition, $ \ccalQ_{\text{goal}}(q_{t})$ includes all states $q$ where the transition from $q_{t}$ to $q$ is an $\epsilon$-transition. Given $ \ccalQ_{\text{goal}}(q_{t})$, the biased action is selected exactly as described in Section \ref{sec:biasedExpl}. Once the set of states $\ccalF_j$ is visited, the set $\ccalV$ is updated as $\ccalV=\ccalV\setminus\{j\}$, and then the set $\ccalQ_{\text{goal}}(q_{t})$ is updated accordingly.


%% file: files/proofs1.tex
\subsection{Proof Proposition \ref{prop:sampleEf1}}
The probability of reaching any state $s_{t+1}=(x_{t+1},q_{t+1})$ where $x_{t+1}\in\ccalX_{\text{closer}}(x_{t})$ under a stochastic policy $\boldsymbol\mu(s,a)$ is: $\sum_{x\in\ccalX_{\text{closer}}}\sum_{a\in\ccalA(x_{t})}\boldsymbol\mu(s_{t},a)P(x_{t},a,x).$
Thus, we have that:\footnote{Note that $q_{t+1}$ is selected deterministically, due to the DRA structure, i.e., $q_{t+1}=\delta_D(q_{t},L(x_{t}))$.}
\begin{align}\label{eq:diff1}
    &\mathbb{P}_b(x_{t+1}\in\ccalX_{\text{closer}}(x_t))- \mathbb{P}_g(x_{t+1}\in\ccalX_{\text{closer}}(x_t))=\nonumber\\&\sum_{x\in\ccalX_{\text{closer}}(x_{t})}\sum_{a\in\ccalA(x_{t})}P(x_{t},a,x)(\boldsymbol\mu_b(s_{t},a)-\boldsymbol\mu_g(s_{t},a)), 
\end{align}
where $\boldsymbol\mu_g$ and $\boldsymbol\mu_b$ refer to the $\epsilon$-greedy (no biased exploration) and $(\epsilon,\delta)$-greedy policy (biased exploration), respectively.
In what follows, we compute $\boldsymbol\mu_b(s_{t},a)-\boldsymbol\mu_g(s_t,a)$, for all $a\in\ccalA_{\mathfrak{P}}$. Recall, that $\boldsymbol\mu_b(s_t,a)$ is the probability of selecting the action $a$ at state $s_t$. Also, hereafter, we assume that the greedy action $a^*$ is different from the biased action $a_b$; however, the same logic applies if $a_b=a^*$, leading to the same result. For simplicity of notation, we use $A=|\ccalA_{\mathfrak{P}}(s)|$.

First, for the action $a=a^*$, we have that (a) $\boldsymbol\mu_b(s_t,a^*)-\boldsymbol\mu_g(s_t,a^*)=(1-\epsilon+\frac{\delta_e}{A})-(1-\epsilon +\frac{\epsilon}{A})=(1-\epsilon+\frac{\delta_e}{A})-(1-\epsilon +\frac{\delta_b+\delta_e}{A})=\frac{\delta_b}{A}$.
%
Similarly, for $a=a_b$, we have that (b) $\boldsymbol\mu_b(s_t,a_b)-\boldsymbol\mu_g(s_t,a_b)=(\delta_b+\frac{\delta_e}{A})-\frac{\epsilon}{A}=-\frac{\delta_b(m-1)}{A}.$
Also, for all other actions $a\neq a_b,a^*$, we have that (c) $\boldsymbol\mu_b(s_t,a)-\boldsymbol\mu_g(s_t,a)=\frac{\delta_e}{A}-\frac{\epsilon}{A}=-\frac{\delta_b}{A}$.
Substituting the above equations (a)-(c) into \eqref{eq:diff1} yields: $\mathbb{P}_b(x_{t+1}\in\ccalX_{\text{closer}})- \mathbb{P}_g(x_{t+1}\in\ccalX_{\text{closer}})=\delta_b\sum_{x\in\ccalX_{\text{closer}}}(P(x_t,a_b,x)-\sum_{a\in\ccalA(x_t)}\frac{P(x_t,a,x)}{A})$.
Due to \eqref{eq:AsAvP} and that $\delta_b>0$, we conclude that $\mathbb{P}_b(s_{t+1}\in\ccalX_{\text{closer}})- \mathbb{P}_g(s_{t+1}\in\ccalX_{\text{closer}})\geq 0$ completing the proof. 

\vspace{-0.6cm}
\subsection{Proof Of Proposition \ref{prop:sampleEf2}}
\vspace{-0.1cm}
\textcolor{black}{The probability of reaching a state $s_{t+1}$ where $x_{t+1}=x_b$ under a policy $\mu(s,a)$ is: $\sum_{a\in\ccalA(x_{t})}\mu(s_{t},a)P(x_{t},a,x_b).$ 
%
Thus, we have $\mathbb{P}_b(x_{t+1}=x_b)- \mathbb{P}_g(x_{t+1}=x_b)=\sum_{a\in\ccalA(x_{t})}P(x_{t},a,x_b)(\mu_b(s_{t},a)-\mu_g(s_{t},a))$
%
Following the same steps as in the proof of Proposition \ref{prop:sampleEf1}, we get that $\mathbb{P}_b(x_{t+1}=x_b)- \mathbb{P}_g(x_{t+1}=x_b)\geq 0$ if $\delta_b>0$, which holds by assumption, and     $P(x_{t},a_b,x_b)\geq\sum_{a\in\ccalA(x_{t})}\frac{P(x_{t},a,x_b)}{|\ccalA(x_{t})|}$ which holds by definition of $a_b$ in \eqref{eq:a_b}. Specifically, given $x_{t}$ and $x_b$, we have that $P(x_{t},a_b,x_b)\geq P(x_{t},a,x_b)$ for all $a\in\ccalA(x_{t})$ due to  \eqref{eq:a_b}. Thus, $P(x_{t},a_b,x_b)$ must be greater than or equal to the average transition probability over the actions $a$ i.e., $\sum_{a\in\ccalA(x_{t})}\frac{P(x_{t},a,x_b)}{|\ccalA(x_{t})|}$ completing the proof.}

%% file: files/proofs2_v3.tex
\vspace{-0.1cm}
\subsection{Proof of Proposition \ref{prop:resultI}}
\vspace{-0.1cm}

\textcolor{black}{By definition of $R_{j^*}$ and $R_{j}$, we can rewrite the inequality $\mathbb{P}_b(R_{j^*}=1)\geq \max_{j\in\ccalJ}\mathbb{P}_{b}(R_{j}=1)$ as 
\begin{align}\label{eq:showResI}  &\prod_{m=0}^{t^*-1}\sum_{a\in\ccalA(x_{t+m})}\mu_b(s_{t+m},a)P(x_{t+m},a,x_{t+m+1})\geq\\&\max_{j\in\ccalJ}\prod_{m=0}^{t^*-1}\sum_{\bar{a}\in\ccalA(\bar{x}_{t+m})}\mu_b(\bar{s}_{t+m},\bar{a})P(\bar{x}_{t+m},\bar{a},\bar{x}_{t+m+1}).\nonumber
\end{align}
where $s_{t+m}=(x_{t+m},q_t)$, $x_{t+m}=p_{j^*}^t(m+1)$, $\bar{s}_{t+m}=(\bar{x}_{t+1},q_t)$, $\bar{x}_{t+m}=p_{j}^t(m+1)$,    for all $m\in\{0,\dots,t^*-1\}$. Recall that by construction of the paths $p_j^t$, the DRA state will remain equal to $q_t$ as the MDP agent moves along any of the paths $p_j^t$, for all $j\in\ccalJ$; see Remark \ref{rem:shortestPath}.
We will show that \eqref{eq:showResI} holds by contradiction. Assume that there exists at least one path $p_{\bar{j}}^t$, $\bar{j}\in\ccalJ$, that does not satisfy \eqref{eq:showResI}, i.e.,
\begin{align}\label{eq:contr}  &\prod_{m=0}^{t^*-1}\sum_{a\in\ccalA(x_{t+m})}\mu_b(s_{t+m},a)P(x_{t+m},a,x_{t+m+1})<\\&\prod_{m=0}^{t^*-1}\sum_{\bar{a}\in\ccalA(\bar{x}_{t+m})}\mu_b(\bar{s}_{t+m},\bar{a})P(\bar{x}_{t+m},\bar{a},\bar{x}_{t+m+1}),\nonumber
\end{align}
where $\bar{s}_{t+m}$ and $\bar{x}_{t+m}$ are defined as per $p_{\bar{j}}^t$.}

\textcolor{black}{
Next, we assume that $a^*\neq a_b$ and $\bar{a}^*\neq \bar{a}_b$; the same logic applies even if this is not the case leading to the same result. Using \eqref{eq:policy}, we plug the values of $\mu_b(s_{t+m},a)$ and $\mu_b(\bar{s}_{t+m},\bar{a})$ for all $a\in\ccalA(x_{t+m})$ and
 $\bar{a}\in\ccalA(\bar{x}_{t+m})$ in \eqref{eq:contr} which yields:
 \begin{align}\label{eq:contrresI}
&\prod_{m=0}^{t^*-1}\{P(x_{t+m},a_b,x_{t+m+1})(\delta_b+\frac{\delta_e}{|\ccalA(x_{t+m})|})+\nonumber\\&P(x_{t+m},a^*,x_{t+m+1})(1-\epsilon+\frac{\delta_e}{|\ccalA(x_{t+m})|})+\nonumber\\
&\sum_{a\neq a^*,a_b}P(x_{t+m},a,x_{t+m+1})(\frac{\delta_e}{|\ccalA(x_{t+m})|})\}<\nonumber\\
&\prod_{m=0}^{t^*-1}\{P(\bar{x}_{t+m},\bar{a}_b,\bar{x}_{t+m+1})(\delta_b+\frac{\delta_e}{|\ccalA(\bar{x}_{t+m})|})+\nonumber\\&P(\bar{x}_{t+m},\bar{a}^*,\bar{x}_{t+m+1})(1-\epsilon+\frac{\delta_e}{|\ccalA(\bar{x}_{t+m})|})+\nonumber\\
&\sum_{\bar{a}\neq \bar{a}^*,\bar{a}_b}P(\bar{x}_{t+m},\bar{a},\bar{x}_{t+m+1})(\frac{\delta_e}{|\ccalA(\bar{x}_{t+m})|})\}.
\end{align}
In \eqref{eq:contrresI}, $a_b$ and  $\bar{a}_b$ stand for the biased action computed when the PMDP state is $s_{t+m}$ and $\bar{s}_{t+m}$ (using the optimal path $p_{j^*}^{t+m}$, as per \eqref{eq:optUnc}, as discussed in Section \ref{sec:biasedExpl}). The same notation extends to all other actions. The purpose of this notation is only to emphasize that the biased and greedy actions at $s_{t+m}$ and $\bar{s}_{t+m}$ are not necessarily the same.
By rearranging the terms in \eqref{eq:contrresI}, we get the following result 
 \begin{align}\label{eq:contrresI2}
&\prod_{m=0}^{t^*-1}\{P(x_{t+m},a_b,x_{t+m+1})\delta_b+\nonumber\\&P(x_{t+m},a^*,x_{t+m+1})(1-\epsilon)+\frac{\delta_e}{|\ccalA(x_{t+m})|}\}<\nonumber\\
&\prod_{m=0}^{t^*-1}\{P(\bar{x}_{t+m},\bar{a}_b,\bar{x}_{t+m+1})\delta_b+\nonumber\\&P(\bar{x}_{t+m},\bar{a}^*,\bar{x}_{t+m+1})(1-\epsilon)+\frac{\delta_e}{|\ccalA(\bar{x}_{t+m})|}\}.
\end{align}
Due to \eqref{eq:Pj}, \eqref{eq:contrresI2} can expressed as $\beta(p_{j^*}^t)<\beta(p_{\bar{j}}^t)$ which contradicts \eqref{reqResI} completing the proof.\footnote{Notice that $\beta(p_{j}^t)$ is equal to the probability that, starting from $x_t$, the MDP path $p_j^t$, $j\in\ccalJ$, will be generated by the end of the time step $t+t^*$, under the proposed $(\epsilon,\delta)$-greedy policy.}}

\vspace{-0.2cm}
\subsection{Proof of Proposition \ref{thm:resultII}}
\textcolor{black}{This proof follows the same steps as the proof of Proposition \ref{prop:resultI}. 
The inequality $\mathbb{P}_b(R_{j^*}=1)\geq \max_{j\in\ccalJ}\mathbb{P}_{g}(R_{j}=1)$ can re-written as 
%
%
\begin{align}\label{eq:toshowBG}
&\prod_{m=0}^{t^*-1}\left(\sum_{a\in\ccalA(x_{t+m})}\mu_b(s_{t+m},a)P(x_{t+m},a,x_{t+m+1})\right)\geq\\&\max_{j\in\ccalJ}\prod_{m=0}^{t^*-1}\left(\sum_{\bar{a}\in\ccalA(\bar{x}_{t+m})}\mu_g(\bar{s}_{t+m},\bar{a})P(\bar{x}_{t+m},\bar{a},\bar{x}_{t+m+1})\right)\nonumber
\end{align}
where $s_{t+m}=(x_{t+m},q_t)$,  $x_{t+m}=p_{j^*}^t(m+1)$,  $\bar{s}_{t+m}=(\bar{x}_{t+m},q_t)$, and $\bar{x}_{t+m}=p_{j}^t(m+1)$,  for all $m\in\{1,\dots,t^*-1\}$. 
%
We will show this result by contradiction. Assume that there exists at least one path $p_{\bar{j}}^t$, $\bar{j}\in\ccalJ$, that does not satisfy \eqref{eq:toshowBG}, i.e.,
\begin{align}\label{eq:contrBG1}  &\prod_{m=0}^{t^*-1}\left(\sum_{a\in\ccalA(x_{t+m})}\mu_b(s_{t+m},a)P(x_{t+m},a,x_{t+m+1})\right)<\\&\prod_{m=0}^{t^*-1}\left(\sum_{\bar{a}\in\ccalA(\bar{x}_{t+m})}\mu_g(\bar{s}_{t+m},\bar{a})P(\bar{x}_{t+m},a,\bar{x}_{t+m+1})\right),\nonumber
\end{align}
where $\bar{s}_{t+m}$ and $\bar{x}_{t+m}$ are defined as per $p_{\bar{j}}^t$.}

\textcolor{black}{In what follows, we denote by $a^*$ and $a_b$ the greedy and the biased action as per $\mu_b$, and $\bar{a}^*$ the greedy action as per $\mu_g$. We assume that $a^*\neq a_b$; the same logic applies even if this is not the case leading to the same final result. 
We plug the values of $\mu_b(s_{t+m},a)$ and $\mu_g(\bar{s}_{t+m},\bar{a})$ for all $a\in\ccalA(x_{t+m})$ and
 $\bar{a}\in\ccalA(\bar{x}_{t+m})$ in \eqref{eq:contrBG1} yielding:
 \begin{align}\label{eq:contrBG2}
&\prod_{m=0}^{t^*-1}\{P(x_{t+m},a_b,x_{t+m+1})\delta_b+\nonumber\\&P(x_{t+m},a^*,x_{t+m+1})(1-\epsilon)+\frac{\delta_e}{|\ccalA(x_{t+m})|}\}<\nonumber\\
&\prod_{m=0}^{t^*-1}\{P(\bar{x}_{t+m},\bar{a}^*,\bar{x}_{t+m+1})(1-\epsilon)+\frac{\epsilon}{|\ccalA(\bar{x}_{t+m})|}\}
\end{align}
Due to \eqref{eq:Pj} and \eqref{eq:Pjg}, the result in \eqref{eq:contrBG2} is equivalent to  $\beta(p_{j^*}^t)<\eta(p_{\bar{j}}^t)$ which contradicts \eqref{reqResII} completing the proof.\footnote{Notice that $\eta(p_{j}^t)$ is equal to the probability that, starting from $x_t$, the MDP path $p_j^t$, will be generated by the end of the time step $t+t^*$, if the PMDP evolves as per the $\epsilon$-greedy policy.}}

\vspace{-0.3cm}
\subsection{Proof of Proposition \ref{thm:resultIII}}
\textcolor{black}{To show this result, it suffices to show that 
\begin{equation}\label{eq:showThm}
  \mathbb{P}_b(x_{t+t^*}\in\ccalX_{\text{goal}})\geq \mathbb{P}_g(x_{t+t^*}\in\ccalX_{\text{goal}}). 
\end{equation}
The reason is that if at the time step $t+t^*$ an MDP state in $\ccalX_{\text{goal}}$ is reached, then at the next time step $t+t^*+1$, a DRA state in $\ccalQ_{\text{goal}}$ will be reached. Notice that the MDP states in $\ccalX_{\text{goal}}$ can be reached at the time step $t+t^*$ if any of the  MDP paths $p_j^t$, $j\in\ccalJ$, originating at $x_t$, are followed. Let $R_{j}$ be a (Bernoulli) random variable that is true if after $t^*$ time steps (i.e., at the time step $t+t^*)$, a path $p_j^t$, $j\in\ccalJ$, has been generated under a policy $\mu$. Then, \eqref{eq:showThm} can be equivalently expressed as: 
\begin{equation}\label{eq:showThmIII}
  \sum_{j\in\ccalJ}\mathbb{P}_b(R_j=1)\geq  \sum_{j\in\ccalJ}\mathbb{P}_g(R_j=1).
\end{equation}
The rest of the proof follows the same logic as the proof of Proposition \ref{thm:resultII}. First, we can rewrite \eqref{eq:showThmIII} as follows:
\begin{align}\label{eq:showThmIII2}  &\sum_{j\in\ccalJ}\left(\prod_{m=0}^{t^*-1}\left(\sum_{a\in\ccalA(x_{t+m})}\mu_b(s_{t+m},a)P(x_{t+m},a,x_{t+m+1})\right)\right)\geq\\&\sum_{j\in\ccalJ}\left(\prod_{m=0}^{t^*-1}\left(\sum_{\bar{a}\in\ccalA(\bar{x}_{t+m})}\mu_g(\bar{s}_{t+m},\bar{a})P(\bar{x}_{t+m},\bar{a},\bar{x}_{t+m+1})\right)\right).\nonumber
\end{align}
Next, as in the proof of Proposition \ref{thm:resultII}, we show that \eqref{eq:showThmIII2} holds by contradiction. Specifically, assume that \eqref{eq:showThmIII2}  does not hold. Then, after plugging the values of $\mu_b(s_{t+m},a)$ and $\mu_g(\bar{s}_{t+m},\bar{a})$ for all $a\in\ccalA(x_{t+m})$ and $\bar{a}\in\ccalA(\bar{x}_{t+m})$ in \eqref{eq:showThmIII2} and after rearranging the terms, we get that
$\sum_{j\in\ccalJ}\beta(p_{j}^t)< \sum_{j\in\ccalJ}\eta(p_{j}^t)$. This contradicts \eqref{eq:reqResIII} completing the proof.}

%% file: files/decay.tex
In this section, we mathematically define the decay rates used for $\epsilon, \delta_b$, and $\delta_e$ in Section \ref{sec:Sim}. 
The parameter $\epsilon$ evolves over episodes \texttt{epi}, as
$\epsilon(\texttt{epi})=1/(\texttt{epi}^{\alpha})$ where $\alpha$ is selected to be equal to $0.1$ for $\mathfrak{M}_1$ and $\mathfrak{M}_2$ and $0.05$ for $\mathfrak{M}_3$. In `Biased 1', $\delta_b$ and $\delta_b$ evolve over episodes, as 
$\delta_b(\texttt{epi})=(1-\frac{1}{\texttt{epi}^{\beta}})\epsilon(\texttt{epi})$ and $\delta_e(\texttt{epi})=\frac{\epsilon(\texttt{epi})}{\texttt{epi}^{\beta}}$. We select $\beta=0.4$ for 
$\mathfrak{M}_1$ and $\mathfrak{M}_2$ and $\beta=0.15$ for $\mathfrak{M}_3$. Observe that $\delta_b(\texttt{epi})+\delta_e(\texttt{epi})=\epsilon(\texttt{epi})$. To define `Biased 2' and `Biased 3', we need first to define the following function denoted by $g(\texttt{epi})$. If $\texttt{epi}<100$, then $g(\texttt{epi})=1 - 0.9\exp(-A \texttt{epi})$. Otherwise, we have that $g(\texttt{epi})=1 - 0.1\exp(-A \texttt{epi})$ for some $A$.  
Then, we have that $\delta_b(\texttt{epi})$ and $\delta_e(\texttt{epi})$ evolve as $\delta_b(\texttt{epi})=(1-g(\texttt{epi}))\epsilon(\texttt{epi})$ and $\delta_e(\texttt{epi})=g(\texttt{epi})\epsilon(\texttt{epi})$.
This choice prioritizes random exploration during the first $100$ episodes.
The larger the $A$, the faster $\delta_b$ converges to $0$. Regarding $\mathfrak{M}_1$ and $\mathfrak{M}_2$, we select $A=0.00015$ for `Biased 2', $A=0.0015$ for `Biased 3', and $A=\infty$ for `Random'.  
As for $\mathfrak{M}_3$, we choose $A = 0.000015$ for `Biased 2' and $A = 0.00015$ for `Biased 3'.

%% file: main.bbl
\begin{thebibliography}{10}
\providecommand{\url}[1]{#1}
\csname url@samestyle\endcsname
\providecommand{\newblock}{\relax}
\providecommand{\bibinfo}[2]{#2}
\providecommand{\BIBentrySTDinterwordspacing}{\spaceskip=0pt\relax}
\providecommand{\BIBentryALTinterwordstretchfactor}{4}
\providecommand{\BIBentryALTinterwordspacing}{\spaceskip=\fontdimen2\font plus
\BIBentryALTinterwordstretchfactor\fontdimen3\font minus \fontdimen4\font\relax}
\providecommand{\BIBforeignlanguage}[2]{{%
\expandafter\ifx\csname l@#1\endcsname\relax
\typeout{** WARNING: IEEEtran.bst: No hyphenation pattern has been}%
\typeout{** loaded for the language `#1'. Using the pattern for}%
\typeout{** the default language instead.}%
\else
\language=\csname l@#1\endcsname
\fi
#2}}
\providecommand{\BIBdecl}{\relax}
\BIBdecl

\bibitem{kiran2021deep}
B.~R. Kiran, I.~Sobh, V.~Talpaert, P.~Mannion, A.~A. Al~Sallab, S.~Yogamani, and P.~P{\'e}rez, ``Deep reinforcement learning for autonomous driving: A survey,'' \emph{IEEE Transactions on Intelligent Transportation Systems}, vol.~23, no.~6, pp. 4909--4926, 2021.

\bibitem{dewey2014reinforcement}
D.~Dewey, ``Reinforcement learning and the reward engineering principle,'' in \emph{AAAI Spring Symposium Series}, 2014.

\bibitem{baier2008principles}
C.~Baier and J.-P. Katoen, \emph{Principles of model checking}.\hskip 1em plus 0.5em minus 0.4em\relax MIT Press, 2008.

\bibitem{wang2015temporal}
J.~Wang, X.~Ding, M.~Lahijanian, I.~C. Paschalidis, and C.~A. Belta, ``Temporal logic motion control using actor--critic methods,'' \emph{The International Journal of Robotics Research}, vol.~34, no.~10, pp. 1329--1344, 2015.

\bibitem{hahn}
E.~M. Hahn, M.~Perez, S.~Schewe, F.~Somenzi, A.~Trivedi, and D.~Wojtczak, ``Omega-regular objectives in model-free reinforcement learning,'' \emph{International Conference on Tools and Algorithms for the Construction and Analysis of Systems}, 2018.

\bibitem{gao2019reduced}
Q.~Gao, D.~Hajinezhad, Y.~Zhang, Y.~Kantaros, and M.~M. Zavlanos, ``Reduced variance deep reinforcement learning with temporal logic specifications,'' in \emph{ACM/IEEE International Conference on Cyber-Physical Systems}, Montreal, Canada, 2019.

\bibitem{bouton2019reinforcement}
M.~Bouton, J.~Karlsson, A.~Nakhaei, K.~Fujimura, M.~J. Kochenderfer, and J.~Tumova, ``Reinforcement learning with probabilistic guarantees for autonomous driving,'' \emph{arXiv preprint arXiv:1904.07189}, 2019.

\bibitem{hasanbeig2019reinforcement}
M.~Hasanbeig, Y.~Kantaros, A.~Abate, D.~Kroening, G.~J. Pappas, and I.~Lee, ``Reinforcement learning for temporal logic control synthesis with probabilistic satisfaction guarantees,'' in \emph{2019 IEEE 58th Conference on Decision and Control (CDC)}, Nice, France, 2019, pp. 5338--5343.

\bibitem{bozkurt2020control}
A.~K. Bozkurt, Y.~Wang, M.~M. Zavlanos, and M.~Pajic, ``Control synthesis from linear temporal logic specifications using model-free reinforcement learning,'' in \emph{2020 IEEE International Conference on Robotics and Automation (ICRA)}, 2020, pp. 10\,349--10\,355.

\bibitem{cai2020learning}
M.~Cai, H.~Peng, Z.~Li, and Z.~Kan, ``Learning-based probabilistic ltl motion planning with environment and motion uncertainties,'' \emph{IEEE Transactions on Automatic Control}, vol.~66, no.~5, pp. 2386--2392, 2020.

\bibitem{lavaei2020formal}
A.~Lavaei, F.~Somenzi, S.~Soudjani, A.~Trivedi, and M.~Zamani, ``Formal controller synthesis for continuous-space mdps via model-free reinforcement learning,'' in \emph{ACM/IEEE 11th International Conference on Cyber-Physical Systems (ICCPS)}.\hskip 1em plus 0.5em minus 0.4em\relax IEEE, 2020, pp. 98--107.

\bibitem{jothimurugan2021compositional}
K.~Jothimurugan, S.~Bansal, O.~Bastani, and R.~Alur, ``Compositional reinforcement learning from logical specifications,'' in \emph{Thirty-Fifth Conference on Neural Information Processing Systems}, 2021.

\bibitem{hasanbeig2022lcrl}
M.~Hasanbeig, D.~Kroening, and A.~Abate, ``Lcrl: Certified policy synthesis via logically-constrained reinforcement learning,'' in \emph{Quantitative Evaluation of Systems: 19th International Conference, QEST 2022, Warsaw, Poland, September 12--16, 2022, Proceedings}.\hskip 1em plus 0.5em minus 0.4em\relax Springer, 2022, pp. 217--231.

\bibitem{hasanbeig2023certified}
H.~Hasanbeig, D.~Kroening, and A.~Abate, ``Certified reinforcement learning with logic guidance,'' \emph{Artificial Intelligence}, vol. 322, 2023.

\bibitem{bozkurt2024learning}
A.~K. Bozkurt, Y.~Wang, M.~M. Zavlanos, and M.~Pajic, ``Learning optimal strategies for temporal tasks in stochastic games,'' \emph{IEEE Transactions on Automatic Control}, 2024.

\bibitem{xuan2024uniqueness}
Z.~Xuan, A.~K. Bozkurt, M.~Pajic, and Y.~Wang, ``On the uniqueness of solution for the bellman equation of ltl objectives,'' in \emph{Learning for Dynamics and Control}, 2024.

\bibitem{hasanbeig2021deepsynth}
M.~Hasanbeig, N.~Y. Jeppu, A.~Abate, T.~Melham, and D.~Kroening, ``Deepsynth: Automata synthesis for automatic task segmentation in deep reinforcement learning,'' in \emph{Proceedings of the AAAI Conference on Artificial Intelligence}, vol.~35, no.~9, 2021, pp. 7647--7656.

\bibitem{icarte2018using}
R.~T. Icarte, T.~Klassen, R.~Valenzano, and S.~McIlraith, ``Using reward machines for high-level task specification and decomposition in reinforcement learning,'' in \emph{International Conference on Machine Learning}.\hskip 1em plus 0.5em minus 0.4em\relax PMLR, 2018, pp. 2107--2116.

\bibitem{wen2020efficiency}
Z.~Wen, D.~Precup, M.~Ibrahimi, A.~Barreto, B.~Van~Roy, and S.~Singh, ``On efficiency in hierarchical reinforcement learning,'' \emph{Advances in Neural Information Processing Systems}, vol.~33, pp. 6708--6718, 2020.

\bibitem{toro2022reward}
R.~Toro~Icarte, T.~Q. Klassen, R.~Valenzano, and S.~A. McIlraith, ``Reward machines:: Exploiting reward function structure in reinforcement learning,'' \emph{Journal of Artificial Intelligence Research}, vol.~73, pp. 173--208, 2022.

\bibitem{cai2022learning}
M.~Cai, M.~Mann, Z.~Serlin, K.~Leahy, and C.-I. Vasile, ``Learning minimally-violating continuous control for infeasible linear temporal logic specifications,'' in \emph{American Control Conference (ACC)}, 2023, pp. 1446--1452.

\bibitem{balakrishnan2023model}
A.~Balakrishnan, S.~Jak{\v{s}}i{\'c}, E.~A. Aguilar, D.~Ni{\v{c}}kovi{\'c}, and J.~V. Deshmukh, ``Model-free reinforcement learning for spatiotemporal tasks using symbolic automata,'' in \emph{62nd IEEE Conference on Decision and Control (CDC)}, Singapore, 2023, pp. 6834--6840.

\bibitem{pathak2017curiosity}
D.~Pathak, P.~Agrawal, A.~A. Efros, and T.~Darrell, ``Curiosity-driven exploration by self-supervised prediction,'' in \emph{International conference on machine learning}.\hskip 1em plus 0.5em minus 0.4em\relax PMLR, 2017, pp. 2778--2787.

\bibitem{zhai2022computational}
Y.~Zhai, C.~Baek, Z.~Zhou, J.~Jiao, and Y.~Ma, ``Computational benefits of intermediate rewards for goal-reaching policy learning,'' \emph{Journal of Artificial Intelligence Research}, vol.~73, pp. 847--896, 2022.

\bibitem{cai2022overcoming}
M.~Cai, E.~Aasi, C.~Belta, and C.-I. Vasile, ``Overcoming exploration: Deep reinforcement learning for continuous control in cluttered environments from temporal logic specifications,'' \emph{IEEE Robotics and Automation Letters}, vol.~8, no.~4, pp. 2158--2165, 2023.

\bibitem{kaelbling1996reinforcement}
L.~P. Kaelbling, M.~L. Littman, and A.~W. Moore, ``Reinforcement learning: A survey,'' \emph{Journal of artificial intelligence research}, vol.~4, pp. 237--285, 1996.

\bibitem{cesa2017boltzmann}
N.~Cesa-Bianchi, C.~Gentile, G.~Lugosi, and G.~Neu, ``Boltzmann exploration done right,'' \emph{Advances in neural information processing systems}, vol.~30, 2017.

\bibitem{chen2017ucb}
R.~Y. Chen, S.~Sidor, P.~Abbeel, and J.~Schulman, ``Ucb exploration via q-ensembles,'' \emph{arXiv preprint arXiv:1706.01502}, 2017.

\bibitem{amin2021survey}
S.~Amin, M.~Gomrokchi, H.~Satija, H.~van Hoof, and D.~Precup, ``A survey of exploration methods in reinforcement learning,'' \emph{arXiv preprint arXiv:2109.00157}, 2021.

\bibitem{fu2014probably}
J.~Fu and U.~Topcu, ``Probably approximately correct {MDP} learning and control with temporal logic constraints,'' \emph{arXiv preprint arXiv:1404.7073}, 2014.

\bibitem{brazdil2014verification}
T.~Br{\'a}zdil, K.~Chatterjee, M.~Chmelik, V.~Forejt, J.~K{\v{r}}et{\'\i}nsk{\`y}, M.~Kwiatkowska, D.~Parker, and M.~Ujma, ``Verification of markov decision processes using learning algorithms,'' in \emph{Automated Technology for Verification and Analysis: 12th International Symposium, ATVA 2014, Sydney, NSW, Australia, November 3-7, 2014, Proceedings 12}.\hskip 1em plus 0.5em minus 0.4em\relax Springer, 2014, pp. 98--114.

\bibitem{fernandez2010probabilistic}
F.~Fern{\'a}ndez, J.~Garc{\'\i}a, and M.~Veloso, ``Probabilistic policy reuse for inter-task transfer learning,'' \emph{Robotics and Autonomous Systems}, vol.~58, no.~7, pp. 866--871, 2010.

\bibitem{kantaros2020stylus}
Y.~Kantaros and M.~M. Zavlanos, ``Stylus*: A temporal logic optimal control synthesis algorithm for large-scale multi-robot systems,'' \emph{The International Journal of Robotics Research}, vol.~39, no.~7, pp. 812--836, 2020.

\bibitem{kantaros2022perception}
Y.~Kantaros, S.~Kalluraya, Q.~Jin, and G.~J. Pappas, ``Perception-based temporal logic planning in uncertain semantic maps,'' \emph{IEEE Transactions on Robotics}, 2022.

\bibitem{ding2014ltl}
X.~Ding, M.~Lazar, and C.~Belta, ``Ltl receding horizon control for finite deterministic systems,'' \emph{Automatica}, vol.~50, no.~2, pp. 399--408, 2014.

\bibitem{lacerda2015optimal}
B.~Lacerda, D.~Parker, and N.~Hawes, ``Optimal policy generation for partially satisfiable co-safe ltl specifications.'' in \emph{International Joint Conference on Artificial Intelligenc}, vol.~15.\hskip 1em plus 0.5em minus 0.4em\relax Citeseer, 2015, pp. 1587--1593.

\bibitem{kantaros2022accelerated}
Y.~Kantaros, ``Accelerated reinforcement learning for temporal logic control objectives,'' in \emph{IEEE/RSJ International Conference on Intelligent Robots and Systems}, Kyoto, Japan, October 2022.

\bibitem{sickert}
S.~Sickert, J.~Esparza, S.~Jaax, and J.~K{\v{r}}et{\'\i}nsk{\`y}, ``Limit-deterministic {B{\"u}chi} automata for linear temporal logic,'' in \emph{CAV}, 2016, pp. 312--332.

\bibitem{cai2021optimal}
M.~Cai, S.~Xiao, Z.~Li, and Z.~Kan, ``Optimal probabilistic motion planning with potential infeasible ltl constraints,'' \emph{IEEE Transactions on Automatic Control}, 2021.

\bibitem{codeAccRL}
Software:, \texttt{\url{https://github.com/kantaroslab/AccRL}}.

\bibitem{kloetzer2008fully}
M.~Kloetzer and C.~Belta, ``A fully automated framework for control of linear systems from temporal logic specifications,'' \emph{IEEE Transactions on Automatic Control}, vol.~53, no.~1, pp. 287--297, 2008.

\bibitem{fainekos2009temporal}
G.~E. Fainekos, A.~Girard, H.~Kress-Gazit, and G.~J. Pappas, ``Temporal logic motion planning for dynamic robots,'' \emph{Automatica}, vol.~45, no.~2, pp. 343--352, 2009.

\bibitem{leahy2016persistent}
K.~Leahy, D.~Zhou, C.-I. Vasile, K.~Oikonomopoulos, M.~Schwager, and C.~Belta, ``Persistent surveillance for unmanned aerial vehicles subject to charging and temporal logic constraints,'' \emph{Autonomous Robots}, vol.~40, no.~8, pp. 1363--1378, 2016.

\bibitem{guo2017distributed}
M.~Guo and M.~M. Zavlanos, ``Distributed data gathering with buffer constraints and intermittent communication,'' in \emph{International Conference on Robotics and Automation}, May-June 2017, pp. 279--284.

\bibitem{kantaros2017distributed}
Y.~Kantaros and M.~M. Zavlanos, ``Distributed intermittent connectivity control of mobile robot networks,'' \emph{IEEE Transactions on Automatic Control}, vol.~62, no.~7, pp. 3109--3121, 2017.

\bibitem{fang2022decentralized}
J.~Fang, Z.~Zhang, and R.~V. Cowlagi, ``Decentralized route-planning for multi-vehicle teams to satisfy a subclass of linear temporal logic specifications,'' \emph{Automatica}, vol. 140, p. 110228, 2022.

\bibitem{vasile2020reactive}
C.~I. Vasile, X.~Li, and C.~Belta, ``Reactive sampling-based path planning with temporal logic specifications,'' \emph{The International Journal of Robotics Research}, vol.~39, no.~8, pp. 1002--1028, 2020.

\bibitem{ding2011ltl}
X.~C.~D. Ding, S.~L. Smith, C.~Belta, and D.~Rus, ``Ltl control in uncertain environments with probabilistic satisfaction guarantees,'' \emph{IFAC Proceedings Volumes}, vol.~44, no.~1, pp. 3515--3520, 2011.

\bibitem{guo2018probabilistic}
M.~Guo and M.~M. Zavlanos, ``Probabilistic motion planning under temporal tasks and soft constraints,'' \emph{IEEE Transanctions on Automatic Control}, 2018.

\bibitem{puterman}
M.~L. Puterman, \emph{{M}arkov decision processes: {D}iscrete stochastic dynamic programming}.\hskip 1em plus 0.5em minus 0.4em\relax John Wiley \& Sons, 2014.

\bibitem{ding2014optimal}
X.~Ding, S.~L. Smith, C.~Belta, and D.~Rus, ``Optimal control of {Markov} decision processes with linear temporal logic constraints,'' \emph{IEEE Trans. on Automatic Control}, vol.~59, no.~5, pp. 1244--1257, 2014.

\bibitem{rlbook}
R.~S. Sutton and A.~G. Barto, \emph{Reinforcement learning: An introduction}.\hskip 1em plus 0.5em minus 0.4em\relax MIT press, 2018.

\bibitem{reachability_in_hybrid}
A.~Abate, M.~Prandini, J.~Lygeros, and S.~Sastry, ``Probabilistic reachability and safety for controlled discrete time stochastic hybrid systems,'' \emph{Automatica}, vol.~44, no.~11, pp. 2724--2734, 2008.

\bibitem{ltl2dstar}
ltl2dstar, \texttt{\url{https://www.ltl2dstar.de/}}.

\end{thebibliography}
